\documentclass{article}

\usepackage[main,final]{neurips_2026}

\newcommand{\mbf}[1]{\mathbf{#1}}

\newcommand{\bF}{\mbf{F}}

\newcommand{\bu}{\mbf{u}}
\newcommand{\bU}{\mbf{U}}
\newcommand{\bv}{\mbf{v}}

\newcommand{\bx}{\mbf{x}}
\newcommand{\bX}{\mbf{X}}

\newcommand{\cf}{\mathrm{CF}}

\newcommand{\cA}{\mathcal{A}}

\newcommand{\cM}{\mathcal{M}}

\newcommand{\iid}{i.\@i.\@d.\ }

\DeclareFontFamily{U}{matha}{\hyphenchar\font45}
\DeclareFontShape{U}{matha}{m}{n}{
      <5> <6> <7> <8> <9> <10> gen * matha
      <10.95> matha10 <12> <14.4> <17.28> <20.74> <24.88> matha12
      }{}
\DeclareSymbolFont{matha}{U}{matha}{m}{n}
\DeclareFontSubstitution{U}{matha}{m}{n}

\DeclareFontFamily{U}{mathx}{\hyphenchar\font45}
\DeclareFontShape{U}{mathx}{m}{n}{
      <5> <6> <7> <8> <9> <10>
      <10.95> <12> <14.4> <17.28> <20.74> <24.88>
      mathx10
      }{}
\DeclareSymbolFont{mathx}{U}{mathx}{m}{n}
\DeclareFontSubstitution{U}{mathx}{m}{n}

\DeclareMathDelimiter{\vvvert}{0}{matha}{"7E}{mathx}{"17}

 \usepackage[utf8]{inputenc} 
\usepackage[T1]{fontenc}    

\usepackage{xcolor}         

\usepackage{url, thmtools}            
\usepackage{booktabs}       
\usepackage{amsfonts}       
\usepackage{nicefrac}       
\usepackage{microtype}      

\usepackage{amsmath,amssymb}
\usepackage{tikz}
\usetikzlibrary{arrows.meta, positioning}
\usepackage{natbib}
\usepackage{subcaption}
\usepackage{amsmath}
\usepackage{amssymb}
\usepackage{mathtools}
\usepackage{amsthm}
\usepackage{algorithm} 
\usepackage{algorithmic} 
\usepackage{enumitem}
\usepackage[T1]{fontenc}

\usepackage{hyperref}

\definecolor{light-gray}{gray}{0.9}
\definecolor{darkblue}{rgb}{0.0,0.0,0.65}
\definecolor{darkred}{rgb}{0.4,0.0,0.0}
\definecolor{mygreen}{RGB}{0,100,0}

\hypersetup{
   colorlinks = true,
   citecolor  = darkblue,
   linkcolor  = darkred,
   filecolor  = darkblue,
   urlcolor   = darkblue,
 }

\theoremstyle{plain}
\newtheorem{theorem}{Theorem}
\newtheorem{proposition}{Proposition}
\newtheorem{lemma}{Lemma}

\theoremstyle{definition}
\newtheorem{definition}{Definition}

\theoremstyle{remark}
\newtheorem{remark}{Remark}

\usepackage[capitalize,nameinlink]{cleveref}

\title{Transporting the Past: An Optimal Transport \\ View of Backtracking Counterfactuals}

\author{$^{1\,}$Zhangyi Liu\thanks{Authors contributed equally.}\, \thanks{This work was carried out while Zhangyi Liu was an intern at the Technical University of Munich.}
  \qquad 
  $^{2,3\,}$Pouria Fatemi\footnotemark[1]
  \qquad
  $^{2,3\,}$Suvrit Sra \\[1em]
  $^{1}\,$Stanford University, Management Science \& Engineering\\[0.1em]
  $^{2}\,$Technical University of Munich, School of CIT, Dept. of Mathematics\\[0.1em]
  $^{3}\,$Munich Center for Machine Learning
}

\begin{document}

\maketitle

\begin{abstract}

We study backtracking counterfactual inference for \emph{Structural Causal Models}~(SCMs) through the lens of optimal transport. In SCMs, a factual observation generally induces a posterior distribution over exogenous variables, making counterfactual inference inherently distributional. We introduce two complementary frameworks that transport this posterior onto the set of exogenous configurations satisfying a desired counterfactual query. The first constructs a natural pointwise projection and is suitable when the transported exogenous distribution can be used directly. The second enforces mutual independence among counterfactual exogenous variables, so that the transported variables remain interpretable as genuine exogenous noises of the same SCM. We show that both frameworks are consistent with the original backtracking counterfactual principle, and we propose marginal, penalty-based, and hard-constrained solvers for finite-posterior settings. Synthetic experiments on non-bijective SCMs illustrate the geometry of the problem and reveal trade-offs among transport cost, counterfactual constraint satisfaction, and preservation of exogenous independence.
\end{abstract}

\section{Introduction}

Counterfactual reasoning asks how a realized system would have evolved under alternative circumstances~\citep{pearl2009causality,pearl2016causal}. In Structural Causal Models (SCMs), the classical abduction--action--prediction procedure gives rise to \emph{interventional counterfactuals}: the exogenous background inferred from the factual evidence is retained, while the structural assignments of the intervened variables are replaced by constants. This semantics is appropriate for questions about the consequences of hypothetical manipulations. Other questions concern the background conditions under which a different event could have occurred \emph{without changing the system's mechanisms}. For such questions, an intervention on a non-root variable severs its ordinary dependence on its causes and may fail to express the intended alternative history. In the terminology of Lewis~\citep{lewis1979counterfactual}, the intervention can be understood as permitting a local ``small miracle.''

Backtracking counterfactuals offer a complementary, mechanism-preserving semantics. They retain the structural assignments and ask which alternative exogenous conditions could have generated a specified counterfactual antecedent~\citep{vonkugelgen2023backtracking,vandenburgh2023backtracking}. The distinction matters because intervening on an event and explaining how that event could have differed need not produce the same answers. Consider, for example, a simplified industrial safety system with a common stop signal $S$ and two production-line stop indicators $A$ and $B$. Suppose the deterministic operating mechanisms are $A=S$ and $B=S$, and in the factual situation $S=A=B=1$. If we ask, ``Had line $A$ continued operating, would line $B$ also have continued?'', an intervention $\mathrm{do}(A=0)$ leaves $S=1$ and hence $B=1$. Under a backtracking interpretation, $A=0$ must instead arise through a different background condition that makes $S=0$, which also yields $B=0$. Both answers follow from their respective assumptions; the choice depends on whether the question concerns an external manipulation or an alternative history under unchanged mechanisms. Such history-oriented queries are relevant to retrospective explanation and some forms of fairness analysis and counterfactual recourse~\citep{bynum2024paradigm}.

In this work, we study backtracking counterfactual inference in \emph{non-bijective} SCMs. In such models, a factual observation may correspond to multiple exogenous configurations, so abduction yields a posterior distribution over latent causes rather than a single latent state. Backtracking inference must therefore transform an entire distribution over exogenous variables while preserving the original structural mechanisms. 
We formulate this problem as a constrained optimal transport problem in exogenous space and study variants with and without an independence constraint.

\subsection{Our Contributions}

Our contributions are as follows.

\begin{enumerate}[label=(\roman*), itemsep = 0mm, topsep = 0mm, leftmargin = 8mm]
    \item \textbf{Framework~(\Cref{sec:framework}).}
    We formulate backtracking counterfactual inference as a constrained optimal transport problem in exogenous space. The proposed framework transports the factual posterior $p(\bU|\bx)$ to the counterfactual space $\cM_{\mathrm{CF}}$ while preserving the original structural assignments. We introduce two complementary variants: \texttt{OT-CF}, which does not impose exogenous independence on the counterfactual distribution, and \texttt{IOT-CF}, which requires the counterfactual exogenous distribution to factorize across coordinates.

    \item \textbf{Theory~(\Cref{sec:analysis_ot_iot}).}
  We provide theoretical support for the transport formulation. In particular, for each fixed factual observation and counterfactual query, we show that our transport solutions can be represented within the probabilistic backtracking framework of \citet{vonkugelgen2023backtracking}. Additionally, we identify conditions under which every optimal \texttt{IOT-CF} solution yields a single counterfactual outcome, illustrating how the independence constraint can restrict the form of optimal counterfactual distributions.

    \item \textbf{Algorithms and experiments~(\Cref{sec:algorithmic-solution-synthetic-experiments}).}
    We develop practical solvers for finite-posterior approximations, including transport solvers for \texttt{OT-CF} and marginal, penalty-based, and hard-constrained variants for \texttt{IOT-CF}. These solvers complement recent optimization-based backtracking methods~\citep{kladny2024deepbacktracking} and explanation-oriented frameworks~\citep{fatemi2025brace}. Controlled experiments examine the solver trade-offs, comparisons with interventional counterfactuals, and sensitivity to the choice of transfer kernel.
\end{enumerate}

\subsection{Motivation and key considerations}

Existing probabilistic approaches describe backtracking through a cross-world transition kernel and conditioning on the counterfactual event~\citep{vonkugelgen2023backtracking,kladny2024deepbacktracking}. This offers a flexible semantics, but the answer depends on the chosen kernel and the corresponding conditional inference may be computationally demanding \citep{fatemi2025brace}. Our sensitivity analysis quantifies how the kernel family and scale affect transport, exogenous dependence, and counterfactual outcome dispersion (\Cref{app:transfer-kernel-sensitivity}). Optimization-based approaches instead select counterfactual latent states according to a similarity or plausibility criterion~\citep{kladny2024deepbacktracking}. Point-valued constructions are particularly convenient when a fully observed factual state identifies a unique exogenous realization, as in invertible SCMs~\citep{pawlowski2020deep,nasresfahany2023bijective}. Our setting allows the factual exogenous state to remain uncertain. Denote the observed factual evidence by $e=(\mathcal O,\bx_{\mathcal O})$, with $\mathcal O\subseteq[n]$, and let $\mu_e=P(\bU\in\cdot\mid\bX_{\mathcal O}=\bx_{\mathcal O})$. For a counterfactual antecedent $\bX_{\cA}^{\mathrm{CF}}=\bx_{\cA}^{\mathrm{CF}}$, feasibility under unchanged mechanisms requires the alternative latent state to lie in
\[
    \cM_{\mathrm{CF}}
    :=\{\bu^{\mathrm{CF}}:\bF_{\cA}(\bu^{\mathrm{CF}})=\bx_{\cA}^{\mathrm{CF}}\}.
\]
The factual posterior $\mu_e$ is therefore the reference distribution, while $\cM_{\mathrm{CF}}$ determines which alternative histories are admissible.
Feasibility alone does not select a counterfactual distribution: many distributions may be supported on $\cM_{\mathrm{CF}}$. We study a \emph{minimal-change} interpretation of backtracking, which prefers alternative background histories requiring the smallest displacement from histories compatible with the factual evidence. Specifically, among all feasible counterfactual exogenous laws $q$, we choose one minimizing the squared $2$-Wasserstein distance,
\[
    q^{\star}\in\arg\min_{q\in\mathcal P_2(\cM_{\mathrm{CF}})}
    W_2^2(\mu_e,q).
\]
Equivalently, this objective minimizes the expected squared distance $\|\bu-\bv\|_2^2$ over couplings between the factual posterior and a feasible counterfactual distribution. It thus determines how posterior mass is paired with admissible histories, rather than identifying one plausible counterfactual point. This notion of similarity is distributional and defined in \emph{exogenous} space, unlike transport methods that operate on observed populations or classifier-induced groups~\citep{villani2009optimal,santambrogio2015optimal,peyre2019computational,charpentier2023optimal,delara2024transport,you2025distributional}.

The transport objective is a modeling choice about which backtracking histories to prefer, rather than a universal consequence of mechanism preservation. It is especially suitable for questions seeking the least revision of the factual background under a specified geometry. Our formulation also covers the limiting case $\mathcal O=\varnothing$, where $\mu_e=P_{\bU}$ and the task becomes a minimal distributional change from the exogenous prior that makes the target event hold under the unchanged structural assignments.

We distinguish two versions of this selection principle. \texttt{OT-CF} (Optimal Transport-Based Counterfactual) imposes feasibility and minimum transport cost, allowing dependence among the counterfactual exogenous coordinates. Such dependence is not in itself incompatible with a Markovian SCM: although its \emph{prior} exogenous distribution factorizes, conditioning on observations or counterfactual events can induce dependence. For applications that require the \emph{counterfactual exogenous law itself} to retain the product form of a Markovian prior, we additionally introduce \texttt{IOT-CF} (Independence-Constrained Optimal Transport Counterfactual), which restricts $q$ to product measures:
\begin{equation*}
   q(\bu^{\mathrm{CF}})
    =
    \prod_i q_i(\bu_i^{\mathrm{CF}}).
\end{equation*}
This requirement preserves the interpretation of the counterfactual exogenous variables as genuine, independently generated noise sources for the original structural mechanisms. The counterfactual condition must then be satisfied without making these noise sources depend on one another. The independence constraint can make the optimization substantially harder~\citep{taskesen2023discrete}. Throughout, we assume the causal graph, structural assignments, and exogenous prior are known. Our concern is counterfactual inference rather than SCM learning or causal discovery. This lets us isolate the geometric and probabilistic issues created by non-bijectivity and exogenous independence.

\subsection{Related Works}

\textbf{Classical counterfactuals and causal explanations.}
Counterfactual reasoning in statistics and causal inference is most commonly grounded in the potential-outcomes framework~\citep{rubin1974estimating,imbens2015causal} or in SCM semantics based on abduction--action--prediction~\citep{pearl2009causality,pearl2016causal,peters2017elements}. These approaches are foundational when the question concerns the effect of an intervention. In machine learning, they also motivate counterfactual explanations and algorithmic recourse methods that search for minimally changed inputs achieving a desired prediction~\citep{wachter2018counterfactual,karimi2021algorithmic,kusner2017counterfactual}. Our setting is different: we assume a known SCM and study mechanism-preserving backtracking counterfactuals rather than nearest observable perturbations or policy interventions.

\textbf{Backtracking counterfactuals.}
Backtracking readings of counterfactuals have a long history in philosophy and logic~\citep{lewis1979counterfactual,hiddleston2005causal,khoo2017backtracking,vandenburgh2023backtracking}. In causal machine learning, \citet{vonkugelgen2023backtracking} provided the central probabilistic SCM formalization of backtracking counterfactuals through a user-specified cross-world kernel on exogenous variables. Building on that formulation, \citet{kladny2024deepbacktracking} developed deep backtracking counterfactuals for causally compliant explanations in deep SCMs. Recent work has also explored natural counterfactuals with only the necessary amount of backtracking~\citep{hao2024natural}, backtracking-based recourse and fairness~\citep{bynum2024paradigm}, and unified explanation frameworks such as BRACE~\citep{fatemi2025brace}. Unlike these methods, we focus on the inference problem induced by non-bijective SCMs, where a factual observation induces a posterior over exogenous states rather than a single latent code.

\textbf{Optimal transport for causal and counterfactual inference.}
Optimal transport provides a natural geometry for comparing distributions with disjoint supports~\citep{villani2009optimal,santambrogio2015optimal,peyre2019computational}. It has recently been used for counterfactual estimation~\citep{charpentier2023optimal}, treatment-effect estimation~\citep{li2023causalot}, transport-based counterfactual models~\citep{delara2024transport}, and distributional counterfactual explanations~\citep{you2025distributional}. These methods typically transport observed distributions, treatment groups, or population-level explanation sets. By contrast, we transport the \emph{posterior over exogenous variables} induced by a factual observation onto a counterfactual space defined by the SCM. This difference is crucial in non-bijective models, where the latent posterior---not the observed sample alone---is the object that must be transformed.

\textbf{Non-bijective SCMs and counterfactual identifiability.}
Recent work notes that tractable counterfactual inference is easiest in invertible or bijective SCMs~\citep{pawlowski2020deep,nasresfahany2023bijective}. More generally, learned SCMs may fail to identify counterfactuals even when interventional behavior is matched, especially outside restrictive invertible settings~\citep{nasr2023counterfactual,chen2025exogenous}. Work on SCM-induced data manifolds also highlights the geometric structure underlying causal and counterfactual reasoning~\citep{dominguezolmedo2023manifolds}. We complement this literature by taking the SCM as given and focusing on the computational and probabilistic consequences of non-bijectivity for backtracking counterfactual inference.

\section{Preliminaries}
\label{sec:preliminaries}

We assume a known Markovian SCM, as defined in \Cref{def:scm}.

\begin{definition}[Markovian SCM; \citealp{pearl2009causality}]\label{def:scm}
Let $G$ be a directed acyclic graph over endogenous variables
$\mathbf X=(X_1,\ldots,X_n)$, and let
$\mathbf U=(U_1,\ldots,U_n)$ be exogenous variables taking values in
$\mathcal U=\prod_{i=1}^n\mathcal U_i$.
A Markovian SCM is a tuple
$\mathcal C=(G,\mathbf X,\mathbf U,\mathbf f,P_{\mathbf U})$, where each
structural assignment
$f_i:\mathcal X_{\mathrm{pa}(i)}\times\mathcal U_i\to\mathcal X_i$
is measurable and
\[
    X_i=f_i(\mathbf X_{\mathrm{pa}(i)},U_i),
    \qquad i=1,\ldots,n.
\]
The exogenous noises are mutually independent, so
$P_{\mathbf U}=\bigotimes_{i=1}^n P_{U_i}$.
Because $G$ is acyclic, the structural assignments induce a measurable
reduced-form map
$F:\mathcal U\to\mathcal X$ satisfying $\mathbf X=F(\mathbf U)$.
The observational distribution is $P_{\mathbf X}=F_{\#}P_{\mathbf U}$.
\end{definition}

Given a factual observation $\mathbf X=\mathbf x$, we denote the abducted
posterior over exogenous variables by
$
    \mu_{\mathbf x}(\mathrm d\mathbf u)
    :=
    P(\mathbf U\in \mathrm d\mathbf u \mid \mathbf X=\mathbf x).
$
A counterfactual query specifies target coordinates
${\mathcal A\subseteq[n]}$ and an alternative value
$\mathbf x_{\mathcal A}^{\mathrm{CF}}$.
Interventional counterfactuals evaluate such a query by the classical
abduction--action--prediction procedure: first infer $\mu_{\mathbf x}$, then
replace the structural assignments of variables in $\mathcal A$ by constants,
and finally propagate the modified SCM forward. Backtracking counterfactuals instead keep all structural assignments fixed and
ask which alternative exogenous histories could have produced the target
condition. Thus the relevant feasible set is the counterfactual space 
    $\mathcal M_{\mathrm{CF}}
    :=
    \left\{
    \mathbf u^{\mathrm{CF}}\in\mathcal U:
    F_{\mathcal A}(\mathbf u^{\mathrm{CF}})
    =
    \mathbf x_{\mathcal A}^{\mathrm{CF}}
    \right\}.$
Interventional counterfactuals make the antecedent true by modifying
mechanisms, whereas backtracking counterfactuals make it true by modifying
exogenous conditions while preserving mechanisms.

Existing probabilistic formulations of backtracking counterfactuals introduce a
cross-world transfer kernel
$K_B(\mathrm d\mathbf u^{\mathrm{CF}}\mid \mathbf u)$ from factual to
counterfactual exogenous variables and then condition on
$\mathbf U^{\mathrm{CF}}\in \mathcal M_{\mathrm{CF}}$
\citep{vonkugelgen2023backtracking}. Backtracking counterfactuals can be
understood as transporting the factual posterior over exogenous histories
through a transition kernel and thereby inducing a distribution over the
counterfactual space. Specifically, given the posterior $\mu_{\mathbf x}$ and a
transition density $k_B$, the induced counterfactual density at a candidate
$\mathbf u^{\mathrm{CF}}\in\mathcal M_{\mathrm{CF}}$ is
\[
    p_B(\mathbf u^{\mathrm{CF}}\mid \mathbf x)
    =
    \int
    k_B(\mathbf u^{\mathrm{CF}}\mid \mathbf u)\,
    \mu_{\mathbf x}(\mathrm d\mathbf u).
\]
This density assigns to each counterfactual exogenous state the average
probability that posterior-compatible factual histories transition to that
state. Common choices such as Gaussian transfer
\[
    k_B(\mathbf u^{\mathrm{CF}}\mid \mathbf u)
    \propto
    \exp\left(
        -\frac{\|\mathbf u^{\mathrm{CF}}-\mathbf u\|_2^2}{2\sigma^2}
    \right)
\]
thus assign larger transition probability to counterfactual histories
closer to factual histories in exogenous space. Subsequent
work~\citep{kladny2024deepbacktracking} returns a single point
counterfactual by selecting the mode of this distribution, yielding a
Maximum A Posteriori (MAP)-type criterion
\[
    \mathbf u^{\mathrm{MAP}}
    \in
    \arg\max_{\mathbf u^{\mathrm{CF}}\in\mathcal M_{\mathrm{CF}}}
    p_B(\mathbf u^{\mathrm{CF}}\mid \mathbf x)
    =
    \arg\max_{\mathbf u^{\mathrm{CF}}\in\mathcal M_{\mathrm{CF}}}
    \int
    k_B(\mathbf u^{\mathrm{CF}}\mid \mathbf u)\,
    \mu_{\mathbf x}(\mathrm d\mathbf u),
\]
thereby reducing point-valued backtracking counterfactual inference to a
maximization problem over the counterfactual feasible set. For more details of
formal definitions and procedures, see
\Cref{app:formal-preliminaries}.

\section{Optimal-Transport Formulation of Backtracking Counterfactuals}
\label{sec:framework}

Building on the probabilistic formulation of backtracking counterfactuals
proposed by \citet{vonkugelgen2023backtracking}, we introduce a complementary
optimal-transport view that is particularly convenient when a factual
observation induces a non-degenerate posterior over exogenous variables. Rather
than specifying a cross-world transition kernel \emph{a priori}, we seek a
counterfactual exogenous distribution that is both feasible under the original
SCM and close to the factual posterior in Wasserstein distance. This perspective
is related in spirit to recent optimization-based approaches to backtracking counterfactuals
\citep{kladny2024deepbacktracking,fatemi2025brace}, as well as to broader
transport-based approaches to causal and counterfactual inference
\citep{charpentier2023optimal,delara2024transport}. 

Using the notation introduced in \Cref{sec:preliminaries}, when
$F_{\mathcal A}$ is sufficiently smooth and
$\mathbf x_{\mathcal A}^{\mathrm{CF}}$ is a regular value,
$\mathcal M_{\mathrm{CF}}$ is a submanifold of the exogenous space
$\mathcal U$. A natural first candidate for comparing the factual posterior and a
counterfactual distribution is the Kullback--Leibler~(KL) divergence. In the
present setting, however, $\mu_{\mathbf x}$ is supported on the factual space
$\{\mathbf u:F(\mathbf u)=\mathbf x\}$, while any counterfactual distribution
must be supported on $\mathcal M_{\mathrm{CF}}$. For nontrivial queries these
supports are typically disjoint, in which case the KL divergence is either
infinite or otherwise uninformative. By contrast, the Wasserstein distance
remains finite on distributions with disjoint supports, provided they have
finite moments, and therefore provides a natural geometry for our problem
\citep{villani2009optimal,santambrogio2015optimal,peyre2019computational}.

With full factual observation and an injective reduced-form map $F$, the posterior is a Dirac measure, and the minimal-change problem reduces to moving one point onto the counterfactual feasible set. Hence, our method also covers the bijective case. When the evidence does not identify a unique latent state, transport instead starts from a non-degenerate posterior. We distinguish \texttt{OT-CF}, which permits arbitrary feasible counterfactual distributions, from \texttt{IOT-CF}, which adds a product-form restriction when required by the intended population-level interpretation of the exogenous law. 

\subsection{Framework \texttt{OT-CF}}
\label{sec:ot-cf}
Let $\mathcal P_2(\mathcal U)$ be the space of Borel probability measures with finite second moments, and assume that $\mathcal U\subseteq\mathbb R^d$ is equipped with the Euclidean ground metric. The \texttt{OT-CF} problem is
\begin{equation}
    q_{\mathrm{OT}}^{\star}
    \in \arg\min_{q\in\mathcal P_2(\cM_{\mathrm{CF}})}
    W_2^2(\mu_e,q),
    \label{eq:ot_cf}
\end{equation}
where the squared transport cost between $\bu$ and $\bv$ is $\|\bu-\bv\|_2^2$. Equivalently, using $\Pi(\mu_e,q)$ for the set of couplings with marginals $\mu_e$ and $q$, the objective minimizes
\[
    \int_{\mathcal U\times\mathcal U}
    \|\bu-\bv\|_2^2\,\gamma(\mathrm d\bu,\mathrm d\bv)
\]
over all couplings $\gamma$ whose first marginal is $\mu_e$ and whose second marginal is supported on $\cM_{\mathrm{CF}}$. Thus, the factual marginal remains fixed while mass is paired with nearby feasible histories. The following result characterizes the optimizer geometrically.

\begin{theorem}[Projection characterization of \texttt{OT-CF}]
\label{thm:ot-cf}
Assume $\cM_{\mathrm{CF}}\subseteq\mathbb R^d$ is nonempty and closed, and $\mu_e\in\mathcal P_2(\mathbb R^d)$. Define the set-valued nearest-point map
\[
    \Gamma(\bu):=\arg\min_{\bv\in\cM_{\mathrm{CF}}}
    \|\bu-\bv\|_2.
\]
If $T:\mathbb R^d\to\cM_{\mathrm{CF}}$ is a measurable selection of $\Gamma$, then $T_{\#}\mu_e$ is an optimizer of \eqref{eq:ot_cf}.
\end{theorem}

Proofs of all theoretical results are provided in \Cref{app:Proofs}. The result in \Cref{thm:ot-cf} should be understood as a geometric characterization rather than a literal closed-form formula. In practice, one still needs to compute or approximate the projection map $T$. Nevertheless, the result clarifies that \texttt{OT-CF} reduces the backtracking problem to projection onto a counterfactual space.

\begin{remark}[Bijective special case]
If $F$ is injective and $\bx$ is fully observed,
then $\mu_e=\delta_{\bu(\bx)}$ for $P_{\mathbf X}$-almost every $\bx$.
An optimal solution to \eqref{eq:ot_cf} is then obtained
by solving the point-to-set problem
\[
    \min_{\bv\in\cM_{\mathrm{CF}}}
    \|\bv-\bu(\bx)\|_2^2
\]
and placing all mass at a minimizer. This recovers the familiar single-latent-state setting of many invertible-model approaches~\citep{kladny2024deepbacktracking,nasresfahany2023bijective}.
However, when the factual vector $\bx$ is only partially observed,
$\mu_e$ need not be a point mass even when $F$ is bijective, so the
distributional formulation \eqref{eq:ot_cf} still applies.
\end{remark}

\subsection{Framework \texttt{IOT-CF}}
\label{sec:iot-cf}

The \texttt{OT-CF} formulation need not produce a product-form counterfactual law. In a Markovian SCM, the \emph{prior} exogenous noises are independent, but conditioning on partial observations or counterfactual events may induce dependence. Such dependence does not, by itself, invalidate backtracking inference. It is nevertheless natural to require the resulting counterfactual exogenous law to retain the product structure of a Markovian prior when it is interpreted as a new population-level noise law. This keeps the counterfactual variables interpretable as genuine, independently generated noise sources for the original structural mechanisms. The counterfactual condition must then be satisfied without relying on dependencies between these sources. To impose this stronger requirement, define
\[
    \mathcal Q_{\mathrm{CF}}^{\otimes}
    :=\left\{q\in\mathcal P_2(\mathcal U):
       q(\cM_{\mathrm{CF}})=1,\quad
       q=\bigotimes_{i=1}^n q_i
       \text{ for some }q_i\in\mathcal P_2(\mathcal U_i)\right\}.
\]
Whenever $\mathcal Q_{\mathrm{CF}}^{\otimes}$ is nonempty, we define
\begin{equation}
    q_{\mathrm{IOT}}^{\star}
    \in\arg\min_{q\in\mathcal Q_{\mathrm{CF}}^{\otimes}}
    W_2^2(\mu_e,q).
    \label{eq:iot_cf}
\end{equation}
Thus, \texttt{IOT-CF} searches for the closest feasible product-form counterfactual exogenous distribution. It alters the exogenous law while keeping structural assignments fixed; it should not be interpreted as retaining every component of the original SCM unchanged.

Relative to \texttt{OT-CF}, the product restriction makes the optimization substantially more difficult. The projection characterization in \Cref{thm:ot-cf} no longer applies directly, because the optimizer must satisfy both geometric feasibility and a nonconvex factorization constraint. This links \eqref{eq:iot_cf} to optimal transport with independent marginals, which can be computationally hard in discrete settings~\citep{taskesen2023discrete}. Accordingly, \texttt{OT-CF} describes a general minimal-change distributional backtracking criterion, while \texttt{IOT-CF} enforces an additional
structural restriction.

\section{Analysis and Properties of \texttt{OT-CF} and \texttt{IOT-CF}}
\label{sec:analysis_ot_iot}

This section studies the mathematical properties of the proposed transport-based
backtracking counterfactuals. We focus on two questions. First, when does the
optimal-transport formulation coincide with probabilistic backtracking
counterfactuals? Second, when does the independence-constrained formulation
have only point-mass optimal solutions?

\subsection{Posterior Factorization under Full Observation}

We first record a basic property of Markovian SCMs. Under
full observation, the posterior over exogenous variables factorizes
coordinate-wise. This justifies treating the factual posterior as a product
source measure in the independence-constrained transport problem.

\begin{proposition}[Posterior factorization under full observation]
\label{prop:posterior-factorization-full}
Consider an acyclic Markovian structural causal model with mutually independent
exogenous noises $P_{\mathbf U}=\bigotimes_{i=1}^n P_{U_i}$ and structural
assignments $X_i=f_i(\mathbf X_{pa(i)},U_i)$. Assume that all endogenous
variables are observed, i.e., $\mathbf X=\mathbf x$, and that the relevant
regular conditional distributions exist. Then, for $P_{\mathbf X}$-almost every
$\mathbf x$,
\begin{equation}
    \mathbb P(\mathbf U\mid \mathbf X=\mathbf x)
    =
    \bigotimes_{i=1}^n
    \mathbb P(U_i\mid X_i=x_i,\mathbf X_{pa(i)}=\mathbf x_{pa(i)}).
    \label{eq:posterior_factorization}
\end{equation}
Equivalently, if
$\mu_{i,\mathbf x}:=\mathbb P(U_i\mid X_i=x_i,\mathbf X_{pa(i)}=\mathbf x_{pa(i)})$,
then $\mu_{\mathbf x}=\bigotimes_{i=1}^n \mu_{i,\mathbf x}$.
\end{proposition}

\subsection{Connection to Probabilistic Backtracking Counterfactuals}
\label{subsec:connection-bcf}

We relate the optimal-transport formulation to the probabilistic backtracking
framework of \citet{vonkugelgen2023backtracking}. In that framework, one specifies a
\emph{cross-world kernel} $P_B(\mathbf U^{\mathrm{CF}}\mid \mathbf U)$, forms the
joint distribution
$P_B(\mathbf U^{\mathrm{CF}},\mathbf U\mid \mathbf x)
=p(\mathbf U\mid \mathbf x)P_B(\mathbf U^{\mathrm{CF}}\mid \mathbf U)$, and then
conditions on the counterfactual feasibility event
$\mathbf F_{\mathcal A}(\mathbf U^{\mathrm{CF}})=\mathbf x^{\mathrm{CF}}_{\mathcal A}$
to obtain the posterior over $\mathbf U^{\mathrm{CF}}$.

Our main observation is that optimal transport already produces a canonical
cross-world joint distribution: an optimal coupling between the factual
posterior and the optimal counterfactual distribution. This coupling induces a
valid backtracking kernel, under which the probabilistic backtracking posterior
recovers exactly the optimal-transport solution.

\begin{theorem}[Compatibility with probabilistic backtracking]
\label{thm:otcf-as-bcf-clean}
Let $q^\star$ be any optimal solution of \eqref{eq:ot_cf} or
\eqref{eq:iot_cf} with finite $r$-th moment, and let $\pi^\star$ be an
optimal transport plan between $p(\cdot\mid\mathbf x)$ and $q^\star$ for the
ground cost $c(\mathbf u,\mathbf u^{\mathrm{CF}})
=\|\mathbf u-\mathbf u^{\mathrm{CF}}\|_r^r$. Then there exists a cross-world
kernel $P_B(\mathbf U^{\mathrm{CF}}\mid \mathbf U)$ such that, under the
probabilistic backtracking construction with factual posterior
$p(\mathbf U\mid\mathbf x)$ and kernel $P_B(\mathbf U^{\mathrm{CF}}\mid\mathbf U)$,
the resulting posterior over counterfactual noises equals $q^\star$:
\begin{equation}
    P_B(\mathbf U^{\mathrm{CF}}\in B\mid
    \mathbf x^{\mathrm{CF}}_{\mathcal A},\mathbf x)
    =q^\star(B),
    \qquad
    B\in\mathcal B(\mathbb R^d).
\end{equation}
\end{theorem}
This is a \emph{fixed-query} representation result. The induced kernel may depend on both the factual observation $\mathbf x$ and the counterfactual query $\mathbf x^{\mathrm{CF}}_{\mathcal A}$. \Cref{thm:otcf-as-bcf-clean} shows that \texttt{OT-CF} and
\texttt{IOT-CF} solutions are always
compatible with the probabilistic backtracking template. Instead of specifying
$P_B(\mathbf U^{\mathrm{CF}}\mid \mathbf U)$ a priori, we induce it from an
optimal transport coupling between the factual posterior and the optimal
counterfactual distribution. This provides an exact non-invertible analogue of
earlier invertible SCM connections that rely on Dirac
posteriors.

We next consider generalized ANMs under full factual observation.
\Cref{prop:3} shows that, under suitable injectivity and feasibility
assumptions, every optimal solution of \texttt{IOT-CF}
in \eqref{eq:iot_cf} is a point mass and thus yields a single
counterfactual outcome. This illustrates how the independence
constraint can restrict the form of optimal counterfactual distributions.

\begin{proposition}
\label{prop:3}
For a generalized ANM
$X_i=g_i(h_i(\mathbf X_{\mathrm{pa}(i)})+k_i(U_i))$ with
invertible $g_i,k_i$ and each $h_i$ jointly injective in its
entire parent vector, suppose $\mathbf X=\mathbf x$ is
fully observed and $\mathcal M_{\mathrm{CF}}$
is nonempty and closed. Then, for $P_{\mathbf X}$-almost every
$\mathbf x$, every optimal solution to \eqref{eq:iot_cf} collapses
to a single point and hence induces a single counterfactual outcome
$\mathbf x^{\mathrm{CF}}$.
\end{proposition}

\section{Algorithmic Solution and Synthetic Experiments}
\label{sec:algorithmic-solution-synthetic-experiments}

This section evaluates the two proposed counterfactual-querying formulations:
\texttt{OT-CF} (\Cref{sec:ot-cf}) and
\texttt{IOT-CF} (\Cref{sec:iot-cf}). The main text gives a concise description of the empirical
solvers and summarizes the experimental findings. Full optimization objectives,
implementation details, hyperparameters, illustrative atom-level solutions,
complete SCM forms, and all experimental tables are provided in
\Cref{app:experiment-details}. We also compare with interventional
counterfactuals and evaluate transfer-kernel sensitivity; these additional
results are summarized in \Cref{subsec:additional-comparisons-main}.

\subsection{Empirical Algorithms in Brief}
\label{subsec:empirical-algorithms-brief}

\textbf{\texttt{OT-CF} mode.}
\Cref{thm:ot-cf} motivates a projection interpretation: after abduction,
the posterior distribution of the exogenous noises is transported to the counterfactual space while minimizing squared displacement. In finite-dimensional
implementations, this can be realized by local projection methods, such
as KKT--Newton or augmented Lagrangian updates. Since \texttt{OT-CF} does not impose an independence
constraint, it can move posterior mass pointwise toward the counterfactual space and may
therefore produce a joint noise distribution that is not marginally independent.

\textbf{\textsc{Marginal} mode.}
The \textsc{Marginal} solver is the simplest independence-preserving implementation. It
learns counterfactual atoms separately for each exogenous marginal and then
forms the joint counterfactual distribution as the product of these optimized marginals.
This gives a low-dimensional and stable parameterization that can strictly keep the independence constraint.

\textbf{\textsc{Penalty} mode.}
The \textsc{Penalty} solver optimizes the full joint support of the abducted posterior.
To avoid producing an arbitrary non-factorized point cloud, it adds an
independence regularizer that encourages particles sharing the same marginal
index to keep the same optimized coordinate. This makes the method more
expressive than the \textsc{Marginal} mode, while still biasing the solution toward a
recombinable product-marginal structure.

\textbf{\textsc{Hard} mode.}
The \textsc{Hard} solver enforces the counterfactual query exactly on every joint support point. It
optimizes only the exogenous variables that are ancestors of the queried
endogenous variables and keeps irrelevant noise coordinates fixed. When a
feasible solution is found, the resulting distribution has zero constraint violation and preserves the intended marginal product structure.

\textbf{Computational scope and scalability.}
Our solvers exploit structure in finite-posterior settings:
\texttt{OT-CF} decomposes into independent posterior-atom projections and is
parallelizable, while ancestor restriction and caching reduce
redundant computation. For \texttt{IOT-CF}, \textsc{Marginal} preserves product
structure by construction and \textsc{Penalty} encourages it through
regularization; both relax counterfactual feasibility. However, if each
structural equation admits $K_i$ compatible roots under full factual observation,
the explicit joint posterior may contain $M=\prod_{i=1}^{d}K_i$ atoms,
potentially exponential in dimension
(\Cref{app:general-experiment-setting}). Marginal parameterization does
not eliminate this joint-support bottleneck in constraint evaluation.
Moreover, exact \texttt{IOT-CF} optimization must combine counterfactual
feasibility with a nonconvex product-measure constraint. Nonsmooth mechanisms
can further hinder local optimization. Our algorithms therefore target
structured finite-posterior instances; scalable inference in much
higher-dimensional SCMs remains open.
\begin{figure*}[t]
  \centering
  \begin{subfigure}[t]{0.49\textwidth}
    \centering
    \includegraphics[width=\linewidth]{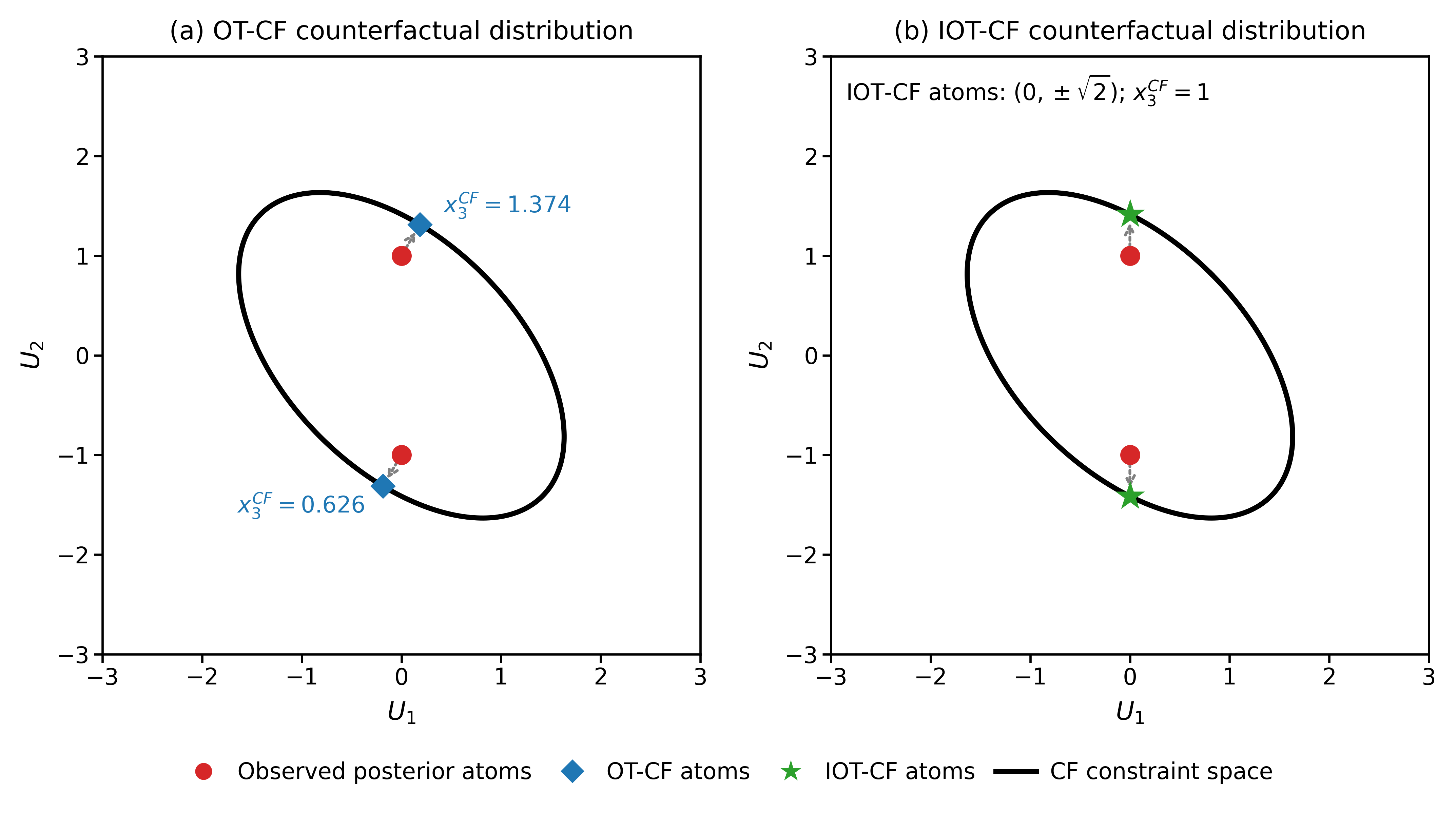}
    \caption{Example~1: quadratic SCM.}
    \label{fig:illustrative-example1}
  \end{subfigure}\hfill
  \begin{subfigure}[t]{0.49\textwidth}
    \centering
    \includegraphics[width=\linewidth]{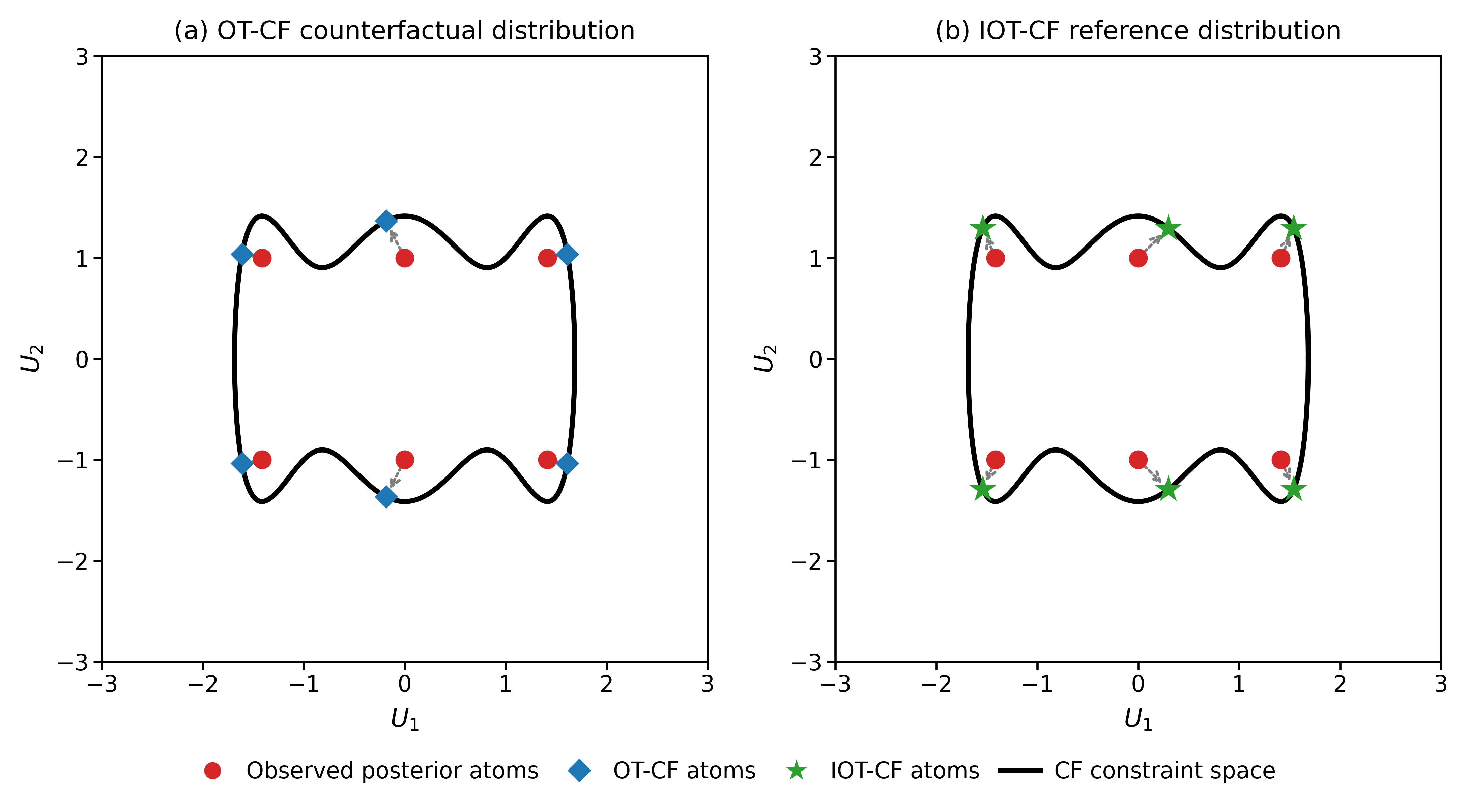}
    \caption{Example~2: multimodal SCM.}
    \label{fig:illustrative-example2}
  \end{subfigure}

  \caption{
  Illustrative backtracking counterfactuals in exogenous space. Red points denote
  posterior atoms compatible with the factual observation, blue diamonds denote
  \texttt{OT-CF} atoms, green stars denote \texttt{IOT-CF} atoms, and black curves
  denote the counterfactual constraint set $x_2^{\mathrm{CF}}=2$.
  In \texttt{OT-CF}, posterior atoms are projected pointwise to nearby feasible
  points on the constraint set. In \texttt{IOT-CF}, the transported counterfactual distribution is constrained to preserve exogenous independence.
  }
  \label{fig:illustrative-examples}
\end{figure*}
\subsection{Illustrative Examples}
\label{subsec:simple-illustrations-main}

We first use two different $3$-node SCMs to illustrate the difference between
\texttt{OT-CF} and \texttt{IOT-CF}. \Cref{fig:illustrative-examples}
shows the results of the two illustrative examples. In both examples, the
factual observation induces a non-degenerate posterior over exogenous atoms,
and the counterfactual query requires transporting this posterior onto the
constraint set $x_2^{\mathrm{CF}}=2$.

The two methods differ in how this transport is performed. \texttt{OT-CF}
uses a pointwise projection rule: each posterior atom is moved independently to
its nearest feasible point on the counterfactual constraint set. This directly
implements a minimal-displacement principle in exogenous space, but the
resulting counterfactual joint distribution need not preserve the original
marginal independence structure. In contrast, \texttt{IOT-CF} transports the
posterior while additionally enforcing that the counterfactual exogenous
distribution remains compatible with marginal independence. Thus,
\texttt{OT-CF} emphasizes nearest feasible pointwise projection, whereas
\texttt{IOT-CF} trades off transport cost with the structural requirement that
the counterfactual distribution factorizes across exogenous coordinates.
Additional atom-level solutions and implementation details are deferred to \Cref{app:illustrative-full}.

It is important to note that, as shown in \Cref{fig:illustrative-examples} (a), the independence constraint can change the resulting counterfactual outcome $\mathbf{x}^{\cf}$. In this example, the independence constraint in \texttt{IOT-CF} makes
the final $\mathbf{x}^{\cf}$ unique. By contrast, if this constraint is ignored and \texttt{OT-CF} is used instead, the method assigns equal probability to two distinct counterfactual outcomes. This demonstrates that the independence constraint is not merely a technical addition, and it can affect the final solution. Ignoring it may therefore lead to semantically implausible counterfactual outcomes.

\begin{table*}[t]
\centering
\caption{Synthetic experiments: exogenous transport cost $D_U$, CF-query residual $L_{\mathrm{CF}}$, and independence violation $L_{\mathrm{ind}}$. Full SCM equations and per-instance details are in \Cref{app:synthetic-full}.}
\label{tab:synthetic-summary}
\footnotesize
\begin{tabular}{llrrr}
\toprule
\textbf{Instance / Query} & \textbf{Method} & $D_U$ & $L_{\mathrm{CF}}$ & $L_{\mathrm{ind}}$ \\
\midrule
5D nonlinear, $(x_2,x_4)=(1.5,3.5)$ & \textsc{Marginal} & $0.250646$ & $1.604\times 10^{-7}$ & $0$ \\
 & \textsc{Penalty} & $0.250813$ & $4.119\times 10^{-8}$ & $5.166\times 10^{-13}$ \\
 & \texttt{OT-CF} & $0.240670$ & $1.694\times 10^{-4}$ & $1.137\times 10^{-21}$ \\
 & \textsc{Hard} & $0.250959$ & $1.169\times 10^{-18}$ & $0$ \\
\midrule
6D nonsmooth, $(x_1,x_5)=(2,4)$ & \textsc{Marginal} & $1.479059$ & $6.977\times 10^{-7}$ & $0$ \\
 & \textsc{Penalty} & $1.479755$ & $1.749\times 10^{-7}$ & $8.480\times 10^{-20}$ \\
 & \texttt{OT-CF} & $1.435321$ & $7.449\times 10^{-4}$ & $2.273\times 10^{-18}$ \\
 & \textsc{Hard} & $1.480453$ & $1.209\times 10^{-18}$ & $0$ \\
\midrule
6D polynomial, $(x_3,x_5)=(1,10)$ & \textsc{Marginal} & $0.783031$ & $2.427\times 10^{-7}$ & $0$ \\
 & \textsc{Penalty} & $0.384916$ & $8.635\times 10^{-6}$ & $1.353\times 10^{-3}$ \\
 & \texttt{OT-CF} & $0.286562$ & $8.432\times 10^{-6}$ & $2.669\times 10^{-2}$ \\
 & \textsc{Hard} & $0.714376$ & $2.176\times 10^{-19}$ & $0$ \\
\midrule
10D nonlinear, $x_8=10$ & \textsc{Marginal} & $1.246304$ & $1.239\times 10^{-7}$ & $0$ \\
 & \textsc{Penalty} & $1.250650$ & $9.216\times 10^{-8}$ & $2.264\times 10^{-6}$ \\
 & \texttt{OT-CF} & $0.795014$ & $3.107\times 10^{-5}$ & $7.270\times 10^{-4}$ \\
 & \textsc{Hard} & $0.798030$ & $7.776\times 10^{-22}$ & $0$ \\
\bottomrule
\end{tabular}
\end{table*}

\subsection{Synthetic Experiments}
\label{subsec:synthetic-experiment-main}

We next evaluate the solvers on nonlinear SCMs of increasing complexity,
including nonsmooth transformations, polynomial equations with multiple roots,
and a ten-dimensional nonlinear system. For each instance, we report the
exogenous transport cost $D_U$, the weighted CF residual $L_{\mathrm{CF}}$, and
an independence violation score $L_{\mathrm{ind}}$. \Cref{tab:synthetic-summary} summarizes the results. The full
SCM definitions, observations, and queries are given in
\Cref{app:synthetic-full}. Metrics and solver hyperparameters
are given in \Cref{app:general-experiment-setting,app:finite-support-solvers}, respectively.

Across the synthetic instances, the \textsc{Marginal} and \textsc{Penalty} solvers
consistently achieve small CF residuals, with \textsc{Penalty} often improving the soft
constraint residual by allowing more flexible movement of the joint support.
\texttt{OT-CF} frequently obtains the smallest $D_U$, but this comes with
larger CF residuals or independence violations because particles are projected
independently. \textsc{Hard} gives exact CF satisfaction and zero independence
violation whenever the constrained problem is successfully solved; however, it
can require larger transport cost.

\subsection{Interventional Comparison and Transfer-Kernel Sensitivity}
\label{subsec:additional-comparisons-main}

\textbf{Comparison with interventional counterfactuals.}
We compare \texttt{OT-CF} with interventional counterfactuals on three synthetic
SCMs (\Cref{app:interventional-comparison},
\Cref{tab:interventional-comparison}). On the 5D nonlinear instance,
\texttt{OT-CF} attains $D_U=0.2407$, compared with an interventional
exogenous-cost diagnostic of $0.4216$, a $42.9\%$ reduction in the reported cost.
We separately assess the compatibility of interventional outcomes with the
original local mechanisms: the feasible posterior mass is $100\%$, $88.08\%$,
and $0\%$ for the 5D nonlinear, 6D polynomial, and 10D nonlinear instances,
respectively. These results illustrate that an intervention can produce
outcomes outside the range of an unchanged local mechanism. The
\texttt{OT-CF} formulation restricts the counterfactual law to the original
mechanism-preserving feasible set; the residuals of its numerical
implementation are reported in \Cref{tab:synthetic-summary}.

\textbf{Transfer-kernel sensitivity.}
For the two illustrative SCMs, we hold the model, factual observation, and
counterfactual antecedent fixed within each experiment and vary the family and
scale of the probabilistic transfer kernel
(\Cref{app:transfer-kernel-sensitivity},
\Cref{tab:transfer-kernel-sensitivity}). Increasing the Gaussian kernel
parameter from $\lambda=0.5$ to $\lambda=50$ changes the reported transport
distance $D_K$ from $2.566089$ to $0.151979$ in Example~1 and from $2.212211$
to $0.070130$ in Example~2. Exogenous dependence and counterfactual outcome
dispersion also change, with different trends across the two examples. This
sensitivity motivates selecting the counterfactual law through an explicit
transport objective and then inducing a kernel from an optimal coupling, as
in \Cref{thm:otcf-as-bcf-clean}.

\section{Conclusions}

In this paper, we introduced a novel counterfactual inference method within the framework of non-bijective SCMs based on backtracking counterfactual reasoning~\citep{vonkugelgen2023backtracking}. We proposed two formulations: the optimal transport counterfactual (\texttt{OT-CF}) and the independence-constrained optimal transport counterfactual (\texttt{IOT-CF}). The \texttt{OT-CF} formulation minimizes the Wasserstein distance between the factual posterior and the counterfactual space, offering a natural solution for counterfactual reasoning in non-bijective SCMs. 
The \texttt{IOT-CF} formulation further enforces independence among counterfactual exogenous variables, so that the transported variables remain interpretable as genuine background noises of the same SCM.

We also presented three solvers offering different trade-offs between computational complexity and the ability to satisfy counterfactual constraints. Our synthetic experiments demonstrated the strengths and limitations of each approach, providing insights into the geometry of the counterfactual problem and the interplay between transport cost, constraint satisfaction, and preservation of exogenous independence. Additional interventional comparisons and transfer-kernel ablations illustrate differences in mechanism compatibility and the sensitivity of probabilistic backtracking to kernel choice.

Despite the promising results, our approach also has limitations. First, while \texttt{IOT-CF} preserves independence among the exogenous variables, it remains computationally challenging. Moreover, the current methods are designed for finite-posterior settings, and extending them to handle more general cases remains an open problem. Future work could refine the optimization algorithms to improve scalability and efficiency.
Finally, applying these methods to real-world causal inference tasks could offer practical insights and guide future developments in counterfactual reasoning.

\section*{Acknowledgments}
Pouria Fatemi and Suvrit Sra acknowledge generous support from the Alexander von Humboldt Foundation. Zhangyi Liu acknowledges generous support from the Top Open Plan of Tsinghua University. The authors declare no competing interests.

\bibliographystyle{apalike}
\bibliography{references_arxiv}

\clearpage
\appendix

\section{Formal Preliminaries and Counterfactual Procedures}
\label[appendix]{app:formal-preliminaries}

This appendix gives the formal definitions and procedures underlying the concise discussion in \Cref{sec:preliminaries}.

\subsection{Structural causal models}

\begin{definition}[Factual posterior]
Given a factual observation $\mathbf X=\mathbf x$, the abducted posterior over exogenous variables is
\[
    \mu_{\mathbf x}(\mathrm d\mathbf u)
    :=
    P(\mathrm d\mathbf u\mid \mathbf X=\mathbf x).
\]
Equivalently, $\mu_{\mathbf x}$ is the conditional law of $\mathbf U$ given 
$F(\mathbf U)=\mathbf x$. 
When $F$ is bijective and the factual observation is complete, 
$\mu_{\mathbf x}$ may be a Dirac measure at the unique exogenous state 
$F^{-1}(\mathbf x)$. 
When $F$ is non-bijective, $\mu_{\mathbf x}$ is generally supported on the space 
$F^{-1}(\mathbf x)=\{\mathbf u:F(\mathbf u)=\mathbf x\}$.
\end{definition}

\begin{definition}[Counterfactual space]
Let $A\subseteq[n]$ be the target coordinates and let 
$\mathbf x_A^{\mathrm{CF}}$ be the desired counterfactual value. 
The counterfactual space is
\[
    M_{\mathrm{CF}}(\mathbf x_A^{\mathrm{CF}})
    :=
    \{\mathbf u^{\mathrm{CF}}\in\mathcal U:
    F_A(\mathbf u^{\mathrm{CF}})=\mathbf x_A^{\mathrm{CF}}\},
\]
where $F_A$ denotes the coordinates of $F$ indexed by $A$.
This is the set of exogenous configurations that produce the counterfactual antecedent while preserving all structural assignments of the original SCM.
\end{definition}

\subsection{Interventional counterfactuals}

The standard SCM-based interventional counterfactual is evaluated by abduction, action, and prediction \citep{pearl2009causality,pearl2016causal}.

\paragraph{Procedure: interventional counterfactual.}
Given an SCM $\mathcal C$, a factual observation $\mathbf x$, target coordinates $A$, and an intervention value $\mathbf a$:

\begin{enumerate}
    \item \textbf{Abduction.} Compute the factual posterior 
    $\mu_{\mathbf x}=P(\mathbf U\in\cdot\mid \mathbf X=\mathbf x)$.

    \item \textbf{Action.} Construct the intervened SCM 
    $\mathcal C_{do(\mathbf X_A=\mathbf a)}$ by replacing the structural assignments for variables in $A$:
    \[
        X_i^{do}=
        \begin{cases}
            a_i, & i\in A,\\
            f_i(\mathbf X_{\mathrm{pa}(i)}^{do},U_i), & i\notin A.
        \end{cases}
    \]
    Let $F^{do(A=\mathbf a)}$ denote the corresponding reduced-form map.

    \item \textbf{Prediction.} Push the abducted posterior through the intervened reduced-form map:
    \[
        P_I(\mathbf X^{\mathrm{CF}}\in C
        \mid \mathbf X=\mathbf x,do(\mathbf X_A=\mathbf a))
        =
        \left(F^{do(A=\mathbf a)}\right)_{\#}\mu_{\mathbf x}(C),
    \]
    for any measurable $C\subseteq\mathcal X$.
\end{enumerate}

This construction changes the structural assignments of the intervened variables. 
It is therefore suited to external manipulation queries, but it does not preserve the original generating mechanism for variables in $A$.

\subsection{Probabilistic backtracking counterfactuals}

Backtracking counterfactuals preserve the original SCM and change the exogenous conditions instead \citep{vonkugelgen2023backtracking}. 
A probabilistic formulation specifies a cross-world transition kernel.

\begin{definition}[Backtracking kernel]
A backtracking kernel is a Markov kernel
\[
    K_B(\mathrm d\mathbf u^{\mathrm{CF}}\mid \mathbf u)
\]
from factual exogenous states $\mathbf u$ to counterfactual exogenous states 
$\mathbf u^{\mathrm{CF}}$.
Given the factual posterior $\mu_{\mathbf x}$, the induced marginal law of counterfactual exogenous variables before conditioning on the target is
\[
    \nu_B(\mathrm d\mathbf u^{\mathrm{CF}}\mid\mathbf x)
    :=
    \int K_B(\mathrm d\mathbf u^{\mathrm{CF}}\mid \mathbf u)\,
    \mu_{\mathbf x}(\mathrm d\mathbf u).
\]
Equivalently, the corresponding cross-world joint law is
\[
    \Pi_B(\mathrm d\mathbf u,\mathrm d\mathbf u^{\mathrm{CF}}\mid\mathbf x)
    :=
    \mu_{\mathbf x}(\mathrm d\mathbf u)
    K_B(\mathrm d\mathbf u^{\mathrm{CF}}\mid\mathbf u).
\]
\end{definition}

\begin{definition}[Probabilistic backtracking posterior]
Assume that 
$\nu_B(M_{\mathrm{CF}}(\mathbf x_A^{\mathrm{CF}})\mid\mathbf x)>0$. 
The probabilistic backtracking posterior over counterfactual exogenous variables is
\[
    P_B(\mathbf U^{\mathrm{CF}}\in B
    \mid \mathbf X=\mathbf x,
    \mathbf U^{\mathrm{CF}}\in M_{\mathrm{CF}})
    =
    \frac{
        \nu_B(B\cap M_{\mathrm{CF}}\mid\mathbf x)
    }{
        \nu_B(M_{\mathrm{CF}}\mid\mathbf x)
    },
\]
for any measurable $B\subseteq\mathcal U$. 
The induced counterfactual outcome distribution is then
\[
    P_B(\mathbf X^{\mathrm{CF}}\in C
    \mid \mathbf X=\mathbf x,
    \mathbf U^{\mathrm{CF}}\in M_{\mathrm{CF}})
    =
    F_{\#}
    P_B(\mathbf U^{\mathrm{CF}}\in\cdot
    \mid \mathbf X=\mathbf x,
    \mathbf U^{\mathrm{CF}}\in M_{\mathrm{CF}})(C).
\]
When $F_A(\mathbf U^{\mathrm{CF}})=\mathbf x_A^{\mathrm{CF}}$ is a zero-probability event, the above conditioning should be understood via regular conditional distributions or by conditioning on a measurable target set.
\end{definition}

\paragraph{MAP interpretation of probabilistic backtracking.}
If the kernel admits a density $k_B(\mathbf u^{\mathrm{CF}}\mid\mathbf u)$ and the goal is to output a single counterfactual exogenous state, the probabilistic construction naturally yields the MAP-type objective
\[
    \mathbf u^{\mathrm{MAP}}
    \in
    \arg\max_{\mathbf u^{\mathrm{CF}}\in M_{\mathrm{CF}}}
    \nu_B(\mathbf u^{\mathrm{CF}}\mid\mathbf x)
    =
    \arg\max_{\mathbf u^{\mathrm{CF}}\in M_{\mathrm{CF}}}
    \int k_B(\mathbf u^{\mathrm{CF}}\mid\mathbf u)
    \mu_{\mathbf x}(\mathrm d\mathbf u).
\]
Thus, for each candidate point $\mathbf u^{\mathrm{CF}}$, the criterion aggregates the probability that all factual exogenous states drawn from the observation-induced posterior $\mu_{\mathbf x}$ transition to that same candidate, and then chooses the candidate with the largest total incoming probability. 
This is the sense in which a point-estimate version of probabilistic backtracking computes the probability of moving the factual exogenous observation distribution to a single point and selects the maximizer.

Our work departs from this principle. 
In non-bijective SCMs, the factual observation usually corresponds to a distribution over possible exogenous histories rather than a single history. 
Therefore, collapsing this distribution to one most likely counterfactual point can lose the geometry of the factual posterior. 
We instead seek a counterfactual distribution by moving each factual exogenous state toward its closest feasible counterfactual counterpart.

\section{Omitted Proofs}
\label[appendix]{app:Proofs}
\begin{proof}[Proof of \Cref{thm:ot-cf}]
Fix any feasible measure $q\in \mathcal Q_{\mathrm{CF}}$ and any coupling $\pi\in\Pi(\mu_{\bx},q)$. Since $q(\mathcal M_{\mathrm{CF}})=1$, the second marginal of $\pi$ is supported on $\mathcal M_{\mathrm{CF}}$, and therefore
\[
\int \lVert \bu-\bv\rVert^r\,\pi(d\bu,d\bv)
\ge
\int d(\bu,\mathcal M_{\mathrm{CF}})^r\,\mu_{\bx}(d\bu).
\]
Taking the infimum over all couplings $\pi\in\Pi(\mu_{\bx},q)$ gives
\[
W_r^r(q,\mu_{\bx})
\ge
\int d(\bu,\mathcal M_{\mathrm{CF}})^r\,\mu_{\bx}(d\bu).
\]
Now let $T$ be a measurable nearest-point selection and define the coupling $\pi_T := (\mathrm{id},T)_{\#}\mu_{\bx}$. Its second marginal is $T_{\#}\mu_{\bx}$, which is feasible because $T(\bu)\in\mathcal M_{\mathrm{CF}}$ for all $\bu$. Moreover,
\[
\int \lVert \bu- T(\bu)\rVert^r\,\mu_{\bx}(d\bu)
=
\int d(\bu,\mathcal M_{\mathrm{CF}})^r\,\mu_{\bx}(d\bu).
\]
Hence $\pi_T$ attains the lower bound above, so $T_{\#}\mu_{\bx}$ is optimal. If the nearest point is unique $\mu_{\bx}$-almost surely, then every optimal coupling must coincide with $(\mathrm{id},T)_{\#}\mu_{\bx}$ almost surely, which implies uniqueness of the optimizer.
\end{proof}

\begin{proof}[Proof of \Cref{prop:posterior-factorization-full}]
Because the graph is acyclic, we may index the variables in a topological order.
The joint law of $(\mathbf U,\mathbf X)$ factorizes as
\begin{equation}
    \mathbb P(d\mathbf u,d\mathbf x)
    =
    \prod_{i=1}^n
    \mathbb P_{U_i}(du_i)\,
    \delta_{f_i(\mathbf x_{pa(i)},u_i)}(dx_i),
\end{equation}
where the parent values $\mathbf x_{pa(i)}$ are already determined by earlier
coordinates in the topological order. For each $i$ and fixed parent assignment
$\mathbf x_{pa(i)}$, disintegrate the local joint law of $(U_i,X_i)$ as
\begin{equation}
    \mathbb P_{U_i}(du_i)\,
    \delta_{f_i(\mathbf x_{pa(i)},u_i)}(dx_i)
    =
    \mathbb P_{X_i\mid \mathbf X_{pa(i)}=\mathbf x_{pa(i)}}(dx_i)\,
    \kappa_i(du_i\mid x_i,\mathbf x_{pa(i)}),
\end{equation}
where $\kappa_i(\cdot\mid x_i,\mathbf x_{pa(i)})
=\mathbb P(U_i\in\cdot\mid X_i=x_i,\mathbf X_{pa(i)}=\mathbf x_{pa(i)})$.
Substituting these local disintegrations into the joint law and conditioning on
$\mathbf X=\mathbf x$ yields
$\mathbb P(d\mathbf u\mid \mathbf X=\mathbf x)=\prod_{i=1}^n
\kappa_i(du_i\mid x_i,\mathbf x_{pa(i)})$, which is exactly
\eqref{eq:posterior_factorization}.
\end{proof}

\begin{proof}[Proof of \Cref{thm:otcf-as-bcf-clean}]
Let $\pi^\star$ be a coupling between $\mathbf U$ and
$\mathbf U^{\mathrm{CF}}$ with marginals $p(\cdot\mid\mathbf x)$ and $q^\star$;
that is, $\mathbf U\sim p(\cdot\mid\mathbf x)$,
$\mathbf U^{\mathrm{CF}}\sim q^\star$, and
$(\mathbf U,\mathbf U^{\mathrm{CF}})\sim\pi^\star$. Since $\pi^\star$ is a joint
distribution on $(\mathbf U,\mathbf U^{\mathrm{CF}})$, there exists a regular
conditional distribution of $\mathbf U^{\mathrm{CF}}$ given $\mathbf U$ under
$\pi^\star$, denoted $\pi^\star(\cdot\mid\mathbf U=\mathbf u)$. Define the
backtracking kernel by
\begin{equation}
\label{eq:kernel-induced-by-coupling}
    P_B(\mathbf U^{\mathrm{CF}}\in B\mid \mathbf U=\mathbf u,\mathbf x)
    :=
    \pi^\star(\mathbf U^{\mathrm{CF}}\in B\mid \mathbf U=\mathbf u),
    \qquad B\in\mathcal B(\mathbb R^d).
\end{equation}
By construction, drawing $\mathbf U\sim p(\cdot\mid\mathbf x)$ and then
$\mathbf U^{\mathrm{CF}}\sim P_B(\cdot\mid\mathbf U,\mathbf x)$ generates the
joint law $\pi^\star$.

Now apply the probabilistic backtracking conditioning step. Feasibility enforces
$\operatorname{supp}(q^\star)\subseteq\mathcal M$, hence
$\mathbf U^{\mathrm{CF}}\in\mathcal M$ almost surely under both $q^\star$ and
$\pi^\star$. Therefore conditioning on
$\{\mathbf U^{\mathrm{CF}}\in\mathcal M\}$, equivalently on
$\{\mathbf F_{\mathcal A}(\mathbf U^{\mathrm{CF}})
=\mathbf x^{\mathrm{CF}}_{\mathcal A}\}$, does not change the joint law:
$P_B(\mathbf U,\mathbf U^{\mathrm{CF}}\mid
\mathbf x^{\mathrm{CF}}_{\mathcal A},\mathbf x)=\pi^\star$. Marginalizing out
$\mathbf U$ gives
$P_B(\mathbf U^{\mathrm{CF}}\in B\mid \mathbf x^{\mathrm{CF}}_{\mathcal A},
\mathbf x)=\pi^\star(\mathbb R^d\times B)=q^\star(B)$, because the second
marginal of $\pi^\star$ is $q^\star$.
\end{proof}

\begin{proof}[Proof of \Cref{prop:3}.]
Consider a counterfactual constraint
$\bx_S^{\mathrm{CF}}=\bx_S^*$ and any optimal product law
$q=\bigotimes_i q_i$.
For a constrained variable $X_j$, with $j\in S$, assume for
contradiction that its counterfactual parent vector
$\mathbf X_{\mathrm{pa}(j)}^{\mathrm{CF}}$ is not a point mass.
Because of the invertible structure, we can write
\[
k_j(U_j^{\mathrm{CF}})
=
g_j^{-1}(x_j^*)
-
h_j(\mathbf X_{\mathrm{pa}(j)}^{\mathrm{CF}}).
\]
Independence of the exogenous noises and acyclicity imply that
$U_j^{\mathrm{CF}}$ is independent of
$\mathbf X_{\mathrm{pa}(j)}^{\mathrm{CF}}$.
However, joint injectivity of $h_j$ makes the right-hand side
a nonconstant function of this parent vector. It cannot also
be independent of that vector, as the left-hand side must be.
This contradicts the independence constraint. Therefore, the
parent vector is a point mass, and invertibility of $k_j$
then forces $U_j^{\mathrm{CF}}$ to be a point mass as well.
Repeating this argument for the parents
shows that all noises associated with the queried nodes and
their ancestors are point masses.

It remains to consider the other noises. Let
$R=S\cup\operatorname{An}(S)$.
Under full observation, invertibility gives
$\mu_e=\mu_{\bx}=\delta_{\mathbf u(\bx)}$
for $P_{\mathbf X}$-almost every $\bx$, and hence
\[
W_2^2(\mu_{\bx},q)
=
\sum_i
\int \|v_i-u_i(\bx)\|_2^2\,q_i(dv_i).
\]
For $i\notin R$, the counterfactual constraint does not depend
on $U_i^{\mathrm{CF}}$. Replacing $q_i$ by
$\delta_{u_i(\bx)}$ therefore preserves feasibility and
independence, while strictly decreasing the transport cost
unless $q_i=\delta_{u_i(\bx)}$ already.
Optimality thus forces these remaining noises to be point
masses as well.

Therefore, every optimal $q$ is a point mass, and the distribution
of every counterfactual variable $X_i^{\mathrm{CF}}$ also collapses
to a point mass. Each optimal solution consequently induces a
single counterfactual outcome. An optimizer exists because
$\mathcal M_{\mathrm{CF}}$ is nonempty and closed.
\end{proof}

\subsection{When Does Pointwise Projection Preserve Independence?}

The unconstrained \texttt{OT-CF} map is the pointwise projection onto
$\mathcal M_{\mathrm{CF}}$, which need not preserve independence. We give
symmetry conditions under which it does: posterior ambiguity must come only from
coordinate-wise symmetries of the structural equations, and
$\mathcal M_{\mathrm{CF}}$ must respect the same symmetries.

\begin{lemma}[Two-node space characterization]
\label{lem:two_node_space}
Consider $X=f_1(U_1)$ and $Y=f_2(X,U_2)$. For
$x\in\mathcal X$ and $y\in\mathcal Y$, let
$S_x(y):=\{u_2:f_2(x,u_2)=y\}$. The following are equivalent:
\begin{enumerate}
    \item For all $x,x'\in\mathcal X$ and $y\in\mathcal Y$,
    $f_2(x',u_2)$ is constant over $S_x(y)$.
    \item There exist maps $h:\mathcal U_2\to\mathcal Z$ and
    $\Phi_x:\mathcal Z\to\mathcal Y$ such that
    $f_2(x,u_2)=\Phi_x(h(u_2))$ for all $x,u_2$, and each $\Phi_x$
    is injective on $h(\mathcal U_2)$.
\end{enumerate}
\end{lemma}

\begin{proposition}[Sufficient symmetry condition for projection to preserve independence]
\label{prop:symmetry_projection_independence}
Consider an acyclic Markovian SCM under full factual observation
$\mathbf X=\mathbf x$, with posterior $\mu_{\mathbf x}$. Assume:
\begin{enumerate}
    \item For each $i$, there is a measurable $h_i$ such that
    $X_i=\Phi_i(\mathbf X_{pa(i)},h_i(U_i))$, and
    $z\mapsto\Phi_i(\mathbf x_{pa(i)},z)$ is injective for every fixed
    $\mathbf x_{pa(i)}$.
    \item For each $i$, a finite group $G_i$ acts isometrically on $U_i$,
    preserves $h_i$, i.e. $h_i(g_i u_i)=h_i(u_i)$ for all $g_i\in G_i$,
    and acts freely and transitively on each space $h_i^{-1}(z)$.
    \item The product group $G:=G_1\times\cdots\times G_n$ acts
    coordinate-wise on $\mathbf u=(u_1,\ldots,u_n)$ and preserves the ground
    metric:
    $\|g\mathbf u-g\mathbf v\|_2=\|\mathbf u-\mathbf v\|_2$ for all $g\in G$.
    \item $\mathcal M_{\mathrm{CF}}$ is nonempty and closed, and the metric
    projection $\Pi_{\mathcal M_{\mathrm{CF}}}(\mathbf u)$ is unique for
    $\mu_{\mathbf x}$-a.e. $\mathbf u$.
\end{enumerate}
Then the \texttt{OT-CF} projection
$T(\mathbf u):=\Pi_{\mathcal M_{\mathrm{CF}}}(\mathbf u)$ is equivariant:
\[
    T(g\mathbf u)=gT(\mathbf u),
    \qquad
    \forall g\in G,\quad \mu_{\mathbf x}\text{-a.e. }\mathbf u .
\]
Consequently, under these symmetry conditions, the unconstrained
\texttt{OT-CF} solution already satisfies the independence constraint in
\eqref{eq:iot_cf}.
\end{proposition}
\begin{proof}[Proof of \Cref{lem:two_node_space}]
First assume the representation
$f_2(x,u_2)=\Phi_x(h(u_2))$ with $\Phi_x$ injective. If
$u_2,u_2'\in S_x(y)$, then
\[
    \Phi_x(h(u_2))
    =
    f_2(x,u_2)
    =
    f_2(x,u_2')
    =
    \Phi_x(h(u_2')).
\]
By injectivity of $\Phi_x$, we obtain
\[
    h(u_2)=h(u_2').
\]
Therefore, for any other $x'$,
\[
    f_2(x',u_2)
    =
    \Phi_{x'}(h(u_2))
    =
    \Phi_{x'}(h(u_2'))
    =
    f_2(x',u_2').
\]
Thus $f_2(x',\cdot)$ is constant over $S_x(y)$.

Conversely, assume the space-constancy condition. Define
\[
    h(u_2)
    :=
    \bigl(f_2(x,u_2)\bigr)_{x\in\mathcal X},
\]
i.e. $h(u_2)$ records the entire response profile of $u_2$ across all parent
values. Let $\Phi_x$ be the coordinate projection:
\[
    \Phi_x(h(u_2)):=f_2(x,u_2).
\]
It remains to
show injectivity of $\Phi_x$ on $h(\mathcal U_2)$. Suppose
\[
    \Phi_x(h(u_2))=\Phi_x(h(u_2')).
\]
This means $f_2(x,u_2)=f_2(x,u_2')$, so $u_2,u_2'$ lie in the same space
$S_x(y)$ for $y=f_2(x,u_2)$. By the assumed space-constancy condition,
\[
    f_2(x',u_2)=f_2(x',u_2')
    \qquad
    \text{for all }x'.
\]
Therefore $h(u_2)=h(u_2')$. Hence $\Phi_x$ is injective on the image
$h(\mathcal U_2)$.
\end{proof}
\begin{proof}[Proof of \Cref{prop:symmetry_projection_independence}]
We first show that the reduced-form map $\mathbf F$ depends on the exogenous
variables only through the tuple
\[
    \bigl(h_1(U_1),\dots,h_n(U_n)\bigr).
\]
This follows by induction in a topological order. Suppose the claim
holds for all parents of $X_i$. Then the parent values
$\mathbf X_{pa(i)}$ are functions only of the corresponding $h_j(U_j)$'s. Since
\[
    X_i=\Phi_i(\mathbf X_{pa(i)},h_i(U_i)),
\]
$X_i$ is also a function only of
$\{h_j(U_j):j\le i\}$.

Now let $g=(g_1,\dots,g_n)\in G$. Since each $g_i$ preserves $h_i$, we have
\[
    h_i(g_i u_i)=h_i(u_i)
    \qquad
    \forall i.
\]
Therefore,
\[
    \mathbf F(g\mathbf u)=\mathbf F(\mathbf u).
\]
In particular, if $\mathbf u\in\mathcal M_{\mathrm{CF}}$, then
\[
    \mathbf F_{\mathcal A}(g\mathbf u)
    =
    \mathbf F_{\mathcal A}(\mathbf u)
    =
    \mathbf x_{\mathcal A}^{\mathrm{CF}},
\]
so $g\mathbf u\in\mathcal M_{\mathrm{CF}}$. Hence
\begin{equation}
    g\mathcal M_{\mathrm{CF}}=\mathcal M_{\mathrm{CF}}.
    \label{eq:manifold_invariant}
\end{equation}

Let $T(\mathbf u)$ denote the unique metric projection of $\mathbf u$ onto
$\mathcal M_{\mathrm{CF}}$. Using that $g$ is an isometry and that
$g\mathcal M_{\mathrm{CF}}=\mathcal M_{\mathrm{CF}}$, we obtain
\[
\begin{aligned}
    T(g\mathbf u)
    &=
    \arg\min_{\mathbf v\in\mathcal M_{\mathrm{CF}}}
    \|\mathbf v-g\mathbf u\|_2                                    \\
    &=
    \arg\min_{\mathbf v\in g\mathcal M_{\mathrm{CF}}}
    \|\mathbf v-g\mathbf u\|_2                                    \\
    &=
    g\left(
        \arg\min_{\mathbf w\in\mathcal M_{\mathrm{CF}}}
        \|\mathbf w-\mathbf u\|_2
    \right)                                                       \\
    &=
    gT(\mathbf u).
\end{aligned}
\]
This proves the equivariance identity.

It remains to show that the pushforward factorizes. Fix a base point
$a_i\in h_i^{-1}(v_i(\mathbf x))$ for each coordinate. Because the $G_i$-action
is free and transitive on the space, every $U_i$ in this space can be written
uniquely as
\[
    U_i=G_i^{\mathbf x}a_i
\]
for a random group element $G_i^{\mathbf x}\in G_i$. Since
$\mu_{\mathbf x}=\bigotimes_i\mu_{i,\mathbf x}$, the random group elements
$G_1^{\mathbf x},\dots,G_n^{\mathbf x}$ are independent.

Let
\[
    \mathbf a=(a_1,\dots,a_n),
    \qquad
    \mathbf w=T(\mathbf a).
\]
For any posterior realization
\[
    \mathbf U
    =
    (G_1^{\mathbf x}a_1,\dots,G_n^{\mathbf x}a_n),
\]
equivariance gives
\[
    T(\mathbf U)
    =
    \bigl(
        G_1^{\mathbf x}w_1,\dots,G_n^{\mathbf x}w_n
    \bigr).
\]
Each counterfactual coordinate depends only on the corresponding independent
random group element $G_i^{\mathbf x}$. Therefore the coordinates of
$T(\mathbf U)$ are mutually independent, and
\[
    T_{\#}\mu_{\mathbf x}
    =
    \bigotimes_{i=1}^n
    \mathcal L(G_i^{\mathbf x}w_i).
\]
This proves that $q_{OT}^{\star}=T_{\#}\mu_{\mathbf x}$ factorizes.
\end{proof}

\section{Experiment Details}
\label[appendix]{app:experiment-details}

This appendix contains the details omitted from the main experimental section:
projection solvers for \texttt{OT-CF}, finite-support implementations of the
\texttt{IOT-CF} solvers, preprocessing and evaluation metrics, hyperparameters,
full illustrative examples, backtracking visualizations, complete synthetic
SCM definitions, interventional comparisons, and transfer-kernel sensitivity
analyses.

\subsection{Projection Details for \texttt{OT-CF}}
\label[appendix]{app:ot-projection-details}

For \texttt{OT-CF}, \Cref{thm:ot-cf} motivates an empirical projection
view: the posterior distribution of the noises is transported toward the CF
manifold by minimizing the displacement cost. In finite-dimensional
implementations, this can be handled by standard projection methods.

\paragraph{KKT--Newton method.}
Consider the constrained optimization problem
\[
    \begin{aligned}
        \min_{y} \quad \|y-x\|_2^2
        \qquad \mathrm{s.t.} \quad g(y)=0,
    \end{aligned}
\]
where $g(y)$ denotes the CF constraint. The KKT--Newton method solves the
linear system
\begin{equation}
    \begin{bmatrix}
        I & Dg(y)^\top\\
        Dg(y) & 0
    \end{bmatrix}
    \begin{bmatrix}
        \Delta y \\
        \lambda
    \end{bmatrix}
    =
    \begin{bmatrix}
        x-y \\
        -g(y)
    \end{bmatrix},
\end{equation}
and then updates $y \gets y+\Delta y$.

\paragraph{Augmented Lagrangian method.}
Instead of enforcing $g(y)=0$ as a hard constraint, the augmented Lagrangian
method incorporates it through a penalty term with coefficient $\rho$:
\begin{equation}
   \min_y \max_\lambda
   L_\rho(y,\lambda)
   =
   \|y-x\|_2^2
   +
   \lambda^\top g(y)
   +
   \rho\|g(y)\|^2.
\end{equation}

\subsection{General Experimental Setting}
\label[appendix]{app:general-experiment-setting}

We evaluate all methods on a collection of SCM instances. Each instance is
specified by a directed acyclic structural model
\[
X_i = f_i(\mathrm{Pa}_i, U_i), \qquad i=1,\ldots,d,
\]
an observed factual sample $x^{\mathrm{obs}}$, and a counterfactual query
\[
\mathcal{Q}=\{(j, a_j): X_j=a_j\}.
\]
\paragraph{Factual posterior construction.}
All solver modes use the same factual posterior representation.
Unless otherwise specified, the exogenous priors are independent
standard Gaussians. For a fully observed factual vector
$x^{\mathrm{obs}}$, define
\[
g_{i,x}(u)=
f_i\!\left(x^{\mathrm{obs}}_{\mathrm{pa}(i)},u\right).
\]
By Proposition~1, the posterior factorizes for almost every full
observation:
\[
\mu_x=\bigotimes_{i=1}^{d}\mu_{i,x},
\qquad
\mu_{i,x}=
\mathcal{L}\!\left(U_i\mid
g_{i,x}(U_i)=x_i^{\mathrm{obs}}\right).
\]
Conditional-law versions at the exceptional observations used below
are specified separately.

If all compatible states are isolated simple roots $u_{i,k}$,
the branch-density version has weights
\[
w_{i,k}
=
\frac{p_i(u_{i,k})/|g'_{i,x}(u_{i,k})|}
{\displaystyle
 \sum_{\ell\in\mathcal{K}_i}
 p_i(u_{i,\ell})/|g'_{i,x}(u_{i,\ell})|},
\]
where $\mathcal{K}_i$ indexes all compatible branches and the
denominator is finite and positive. This formula covers finite
simple-root posteriors and normalizable countable simple-root
posteriors. It does not directly cover critical roots or inverse
images of positive prior probability.

\paragraph{Periodic and critical components.}
Gaussian priors do not make periodic inverse images finite.
For the critical constraint $\cos U=1$, we choose the local
shrinking-neighborhood version
\[
\mathcal{L}(U\mid\cos U=1)
:=
\lim_{\epsilon\downarrow0}
\mathcal{L}\!\left(U\mid|\cos U-1|\leq\epsilon\right).
\]
For a standard Gaussian prior, this weak limit is
\[
\sum_{k\in\mathbb{Z}}
\frac{\varphi(2\pi k)}
{\sum_{\ell\in\mathbb{Z}}\varphi(2\pi\ell)}
\,\delta_{2\pi k},
\]
where $\varphi$ is the standard Gaussian density. The local
conditioning intervals have equal widths, giving branch masses
proportional to the prior density without division by a zero
derivative. Parent values are fixed at their factual observations.
At the exceptional full observations, we use the product of these
specified local versions. A global noisy-observation model would
require a separate likelihood and factorization analysis.

\paragraph{Posterior structure of the synthetic instances.}
In the five-dimensional nonlinear instance, the posterior support is
\[
U_1\in\pi\mathbb{Z},\quad U_2\in\{-1,1\},\quad
U_3\in2\pi\mathbb{Z},\quad U_4\in\{-1,1\},\quad U_5=0.
\]
The periodic branch weights are proportional to
$\varphi(k\pi)$ for $U_1$ and $\varphi(2\pi k)$ for $U_3$;
the two signs of $U_2$ and $U_4$ each have probability $1/2$.

In the six-dimensional nonsmooth instance,
$U_1\in\{-2,2\}$ with equal masses, $U_2\in2\pi\mathbb{Z}$
with the critical-level weights above, and $U_5=U_6=0$.
The other two marginals are continuous:
\[
\begin{aligned}
\mu_{3,x}(du)
&=2\varphi(u)\mathbf{1}_{\{u>0\}}\,du,\\
\mu_{4,x}(du)
&=\frac{\varphi(u)\mathbf{1}_{\{|u|\geq2\}}}
        {2[1-\Phi(2)]}\,du,
\end{aligned}
\]
where $\Phi$ is the standard Gaussian cumulative distribution
function. Their numerical representations must approximate these
conditional distributions.

The ten-dimensional instance has countably many inverse branches
for its periodic mechanisms involving $U_2,U_6,U_8,U_{10}$.
The six-dimensional polynomial instance instead has
$(3,4,2,2,2,2)$ simple roots across coordinates, giving $192$
joint atoms when all branches are retained.

\paragraph{Finite representation supplied to the solvers.}
Write the numerical source as
\[
\widehat{\mu}_x
=\bigotimes_{i=1}^{d}\widehat{\mu}_{i,x},
\qquad
\widehat{\mu}_{i,x}
=\sum_{k=1}^{K_i}\widehat w_{i,k}\delta_{u_{i,k}},
\qquad
\sum_{k=1}^{K_i}\widehat w_{i,k}=1.
\]
For finite simple-root posteriors, retaining every branch gives
the exact posterior. For a countably supported marginal, retaining
a finite branch set requires renormalization:
\[
\widehat w_{i,k}=\frac{w_{i,k}}{1-\tau_i},
\]
where $\tau_i$ is the omitted conditional mass. The total mass
omitted by these branch truncations is
$1-\prod_i(1-\tau_i)$, with $\tau_i=0$ for coordinates without
branch truncation. Continuous marginals additionally require a
specified quadrature or sampling rule.

The joint particles and weights are
\[
M=\prod_{i=1}^{d}K_i,\qquad
u_m^0=(u_{1,m_1},\ldots,u_{d,m_d}),\qquad
\pi_m=\prod_{i=1}^{d}\widehat w_{i,m_i}.
\]
All solver modes receive this same finite source. Below, $w_{i,k}$
denotes the normalized numerical weight $\widehat w_{i,k}$.

For every method, the counterfactual residual is evaluated in endogenous space:
\[
R_{\mathrm{CF}}(\tilde u_m)
=
\sum_{(j,a_j)\in\mathcal{Q}}
\big(f_j(\tilde u_m)-a_j\big)^2 .
\]
We report the exogenous transport cost and, when needed, the endogenous
transport cost:
\[
D_U=\sum_{m=1}^{M}\pi_m\|\tilde u_m-u_m^0\|_2^2,
\qquad
D_X=\sum_{m=1}^{M}\pi_m\|f(\tilde u_m)-x^{\mathrm{obs}}\|_2^2 .
\]
We also report the weighted counterfactual constraint violation
$\sum_m \pi_m R_{\mathrm{CF}}(\tilde u_m)$ and the independence violation used
to measure whether the optimized joint particles still respect the original
product-marginal structure.

\subsection{Finite-Support Solver Modes}
\label[appendix]{app:finite-support-solvers}

We compare four solver modes implemented in the codebase.

\paragraph{\textsc{Marginal}.}
This mode optimizes counterfactual marginal atoms $\{\tilde u_{i,k}\}$ while
keeping the product posterior weights fixed. The CF joint distribution is
reconstructed as the product of the optimized marginals:
\begin{equation}
\tilde P^{\mathrm{CF}}(U)
=
\prod_i
\sum_k w_i^{(k)} \delta_{\tilde u_i^{(k)}}.
\end{equation}
The objective is
\begin{equation}
\sum_{i=1}^{d}\sum_{k=1}^{K_i}
w_{i,k}(\tilde u_{i,k}-u_{i,k})^2
+
\lambda
\sum_{m=1}^{M}\pi_m R_{\mathrm{CF}}(\tilde u_m),
\end{equation}
where the optimized joint support $\{\tilde u_m\}$ is reconstructed from the
optimized marginal atoms. This parameterization is simple and stable, but it
enforces the CF query softly through an expected residual.

\paragraph{\textsc{Penalty}.}
This mode directly optimizes the full joint support
$\{\tilde u_m\}_{m=1}^{M}$. Its objective is
\begin{equation}
D_U
+
\lambda_{\mathrm{CF}}\sum_{m=1}^{M}\pi_m R_{\mathrm{CF}}(\tilde u_m)
+
\lambda_{\mathrm{ind}} I(\{\tilde u_m\}),
\end{equation}
where $I(\cdot)$ penalizes violations of the product-marginal structure by
encouraging particles that share the same marginal atom index to share the same
optimized coordinate. Compared with the \textsc{Marginal} mode, \textsc{Penalty} is more
flexible, but the CF query and independence structure are both enforced softly.

\paragraph{Hard.}
This mode solves a hard-constrained optimization over marginal atoms. Only
noise variables that are ancestors of the queried endogenous variables are
optimized. Let $\mathcal{A}(\mathcal{Q})$ denote this ancestor noise set. The
optimization problem is
\begin{equation}
\begin{aligned}
\min_{\{\tilde u_{i,k}: i\in\mathcal{A}(\mathcal{Q})\}}
&\quad
\sum_{i\in\mathcal{A}(\mathcal{Q})}\sum_{k=1}^{K_i}
w_{i,k}(\tilde u_{i,k}-u_{i,k})^2 \\
\mathrm{s.t.}
&\quad
f_j(\tilde u_m)=a_j,
\qquad
\forall (j,a_j)\in\mathcal{Q},\ \forall m .
\end{aligned}
\end{equation}
Each restart first minimizes a quadratic penalty objective and is then refined
with a constrained optimizer.

\paragraph{\texttt{OT-CF}.}
This mode projects each observed joint particle independently. For each
particle $u_m^0$, only the ancestor noise coordinates are optimized:
\begin{equation}
\min_{\tilde u_m}
\|\tilde u_m-u_m^0\|_2^2
+
\rho R_{\mathrm{CF}}(\tilde u_m).
\end{equation}
Repeated ancestor-coordinate configurations are cached, so identical projection
subproblems are solved only once. The final transport cost is computed as the
posterior-weighted average over all projected particles.

\begin{table}[t]
\centering
\caption{Hyperparameter configuration. A dash denotes a parameter not used by that mode. Hard uses $5000$ warm-start updates followed by $20\times1000$ augmented-Lagrangian updates.}
\label{tab:solver-hyperparams-app}
\scriptsize
\setlength{\tabcolsep}{3pt}
\begin{tabular}{@{}lcccc@{}}
\toprule
Parameter & \textsc{Marginal} & \textsc{Penalty} & \textsc{Hard} & \texttt{OT-CF} \\
\midrule
Optimizer & Adam & Adam & Adam & Adam \\
Full-batch updates & $5000$ & $5000$ & $5000+20000$ & $20000$ \\
Initial learning rate & $10^{-3}$ & $10^{-3}$ & $10^{-3}$ & $10^{-3}$ \\
CF penalty / warm-start weight & $1000$ & $2000$ & $30$ & $30$ \\
Independence penalty & -- & $2000$ & -- & -- \\
Restarts & $1$ & $1$ & $10$ & $10$ \\
Random restart standard deviation & -- & -- & $0.5$ & $0.5$ \\
Maximum-query-error tolerance & -- & -- & $10^{-9}$ & -- \\
\bottomrule
\end{tabular}
\end{table}

\subsection{Illustrative Examples: Full Atom-Level Solutions}
\label[appendix]{app:illustrative-full}

\paragraph{Example 1.}
Consider the following three-node SCM:
\[
    \begin{cases}
        X_1 = U_1,\\
        X_2 = X_1^2 + X_1U_2 + U_2^2,\\
        X_3 = X_1X_2 + U_3.
    \end{cases}
\]
We show that the optimal solutions of
\eqref{eq:ot_cf} and \eqref{eq:iot_cf} are different. Assume that the
exogenous noises $U_i$ are independent with distribution
$\mathrm{Unif}[-3,3]$. For the observation $(0,1,1)$, the posterior assigns
equal probability to $(0,1,1)$ and $(0,-1,1)$. Under the CF query $X_2=2$, the
solutions obtained from \eqref{eq:ot_cf} and \eqref{eq:iot_cf} are shown in
\Cref{tab:example1-reference}.

\begin{table}[t]
\centering
\small
\caption{Counterfactual references for Example~1.
OT projection coordinates and outcomes are rounded to six decimals.
The IOT-CF row uses exact coordinates. The third noise coordinate
remains equal to $1$.}
\label{tab:example1-reference}
\begin{tabular}{@{}lccc@{}}
\toprule
Method & Source $(0,1)$ & Source $(0,-1)$
& Counterfactual $X_3$\\
\midrule
OT-CF
& $(0.186826,1.311515)$
& $(-0.186826,-1.311515)$
& $\substack{1.373652,\ 0.626348\\
              \text{with equal probability}}$\\
IOT-CF
& $(0,\sqrt{2})$
& $(0,-\sqrt{2})$
& $1$\\
\bottomrule
\end{tabular}
\end{table}

\paragraph{Example 2.}
Consider the following three-node SCM:
\[
    \begin{cases}
        X_1 = U_1^3 - 2U_1,\\
        X_2 = X_1^2 + U_2^2,\\
        X_3 = X_1 + X_2 + U_3.
    \end{cases}
\]
Take the observation $(X_1,X_2,X_3)=(0,1,1)$ and independent
$\operatorname{Unif}[-3,3]$ priors. Write $h(t)=t^3-2t$.
The factual posterior is supported on
\[
U_1\in\{0,\sqrt{2},-\sqrt{2}\},
\qquad U_2\in\{1,-1\},
\qquad U_3=0.
\]
Under the inverse-branch conditional-density convention,
the marginal weights of $U_1$ are proportional to
$p_1(u_1)/|h'(u_1)|$. Since
$|h'(0)|=2$ and $|h'(\pm\sqrt{2})|=4$, the posterior is
\[
\mu_x=
\left(
 \tfrac12\delta_0+
 \tfrac14\delta_{\sqrt{2}}+
 \tfrac14\delta_{-\sqrt{2}}
\right)
\otimes
\left(\tfrac12\delta_1+\tfrac12\delta_{-1}\right)
\otimes\delta_0.
\]
Thus the joint masses at
$(0,1,0)$, $(0,-1,0)$,
$(\sqrt{2},1,0)$, $(\sqrt{2},-1,0)$,
$(-\sqrt{2},1,0)$, and $(-\sqrt{2},-1,0)$ are
\[
\left(\tfrac14,\tfrac14,\tfrac18,\tfrac18,\tfrac18,\tfrac18\right),
\]
respectively. Table~\ref{tab:example2-reference} provides
numerical projection references and an explicitly feasible
product-law comparison for this posterior.

\paragraph{Projection and feasible product references.}
For the query $X_2^{\mathrm{CF}}=2$, the feasible set satisfies
\[
h(v_1)^2+v_2^2=2.
\]
The third coordinate is optimally left at $v_3=0$.
The OT projection references in
Table~\ref{tab:example2-reference} have posterior-weighted
squared displacement approximately $0.1044628684$.
For each source with $u_1=0$, both signs of the first projected
coordinate give tied nearest points; the table selects the negative
sign.

For an explicitly feasible product law, set
\[
r=1.294,\qquad s=\sqrt{2-r^2}.
\]
Let $a\in(0,\sqrt{2/3})$ satisfy $h(a)=-s$, and let
$b>\sqrt{2}$ satisfy $h(b)=s$. Define
\[
q_{\mathrm{ref}}
=
\left(\tfrac12\delta_a+\tfrac14\delta_b+\tfrac14\delta_{-b}\right)
\otimes\tfrac12(\delta_r+\delta_{-r})\otimes\delta_0.
\]
Every point in its Cartesian support satisfies the query exactly.
The coordinatewise source--target assignments preserve order, so
one-dimensional optimality and tensorization give
\[
W_2^2(\mu_x,q_{\mathrm{ref}})
=
\tfrac12a^2+\tfrac12(b-\sqrt{2})^2+(r-1)^2
\approx0.1388877952.
\]
This construction supplies a feasible comparison law and an upper
bound on the IOT-CF optimum. No global IOT-CF optimality is claimed.

\begin{table}[t]
\centering
\small
\setlength{\tabcolsep}{5pt}
\caption{Reference configurations for Example~2.
The OT column reports numerical nearest-point references.
The product column reports the explicitly constructed feasible law
$q_{\mathrm{ref}}$. All third coordinates are zero.
Coordinates are rounded to six decimals; product-law feasibility
refers to the exact root definitions in the text.
These columns provide validation references rather than outputs
from rerunning the three IOT-CF solvers.}
\label{tab:example2-reference}
\begin{tabular}{@{}cccc@{}}
\toprule
Source $(u_1,u_2)$ & Mass
& OT projection & Feasible product reference\\
\midrule
$(0,1)$ & $1/4$
& $(-0.185354,1.366476)$
& $(0.298603,1.294000)$\\
$(0,-1)$ & $1/4$
& $(-0.185354,-1.366476)$
& $(0.298603,-1.294000)$\\
$(\sqrt{2},1)$ & $1/8$
& $(1.611487,1.036718)$
& $(1.539671,1.294000)$\\
$(\sqrt{2},-1)$ & $1/8$
& $(1.611487,-1.036718)$
& $(1.539671,-1.294000)$\\
$(-\sqrt{2},1)$ & $1/8$
& $(-1.611487,1.036718)$
& $(-1.539671,1.294000)$\\
$(-\sqrt{2},-1)$ & $1/8$
& $(-1.611487,-1.036718)$
& $(-1.539671,-1.294000)$\\
\midrule
\multicolumn{2}{@{}l}{Squared transport cost}
& $0.1044628684$ & $0.1388877952$\\
\bottomrule
\end{tabular}
\end{table}

\subsection{Backtracking Counterfactual Visualizations}
\label[appendix]{app:backtracking-visuals}

\begin{figure*}[htbp]
  \centering
  \begin{subfigure}[t]{0.3\textwidth}
    \centering
    \includegraphics[width=\linewidth]{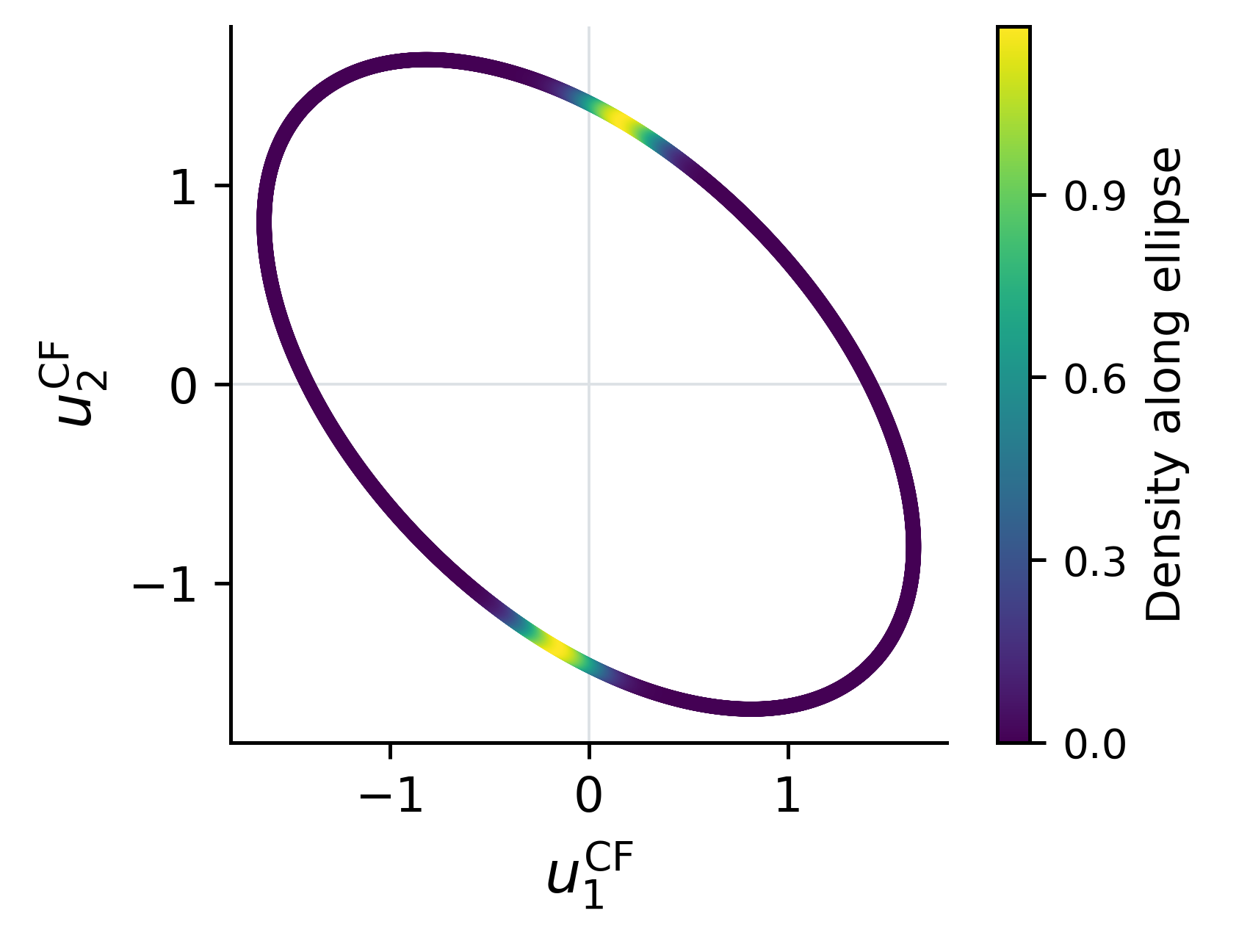}
    \caption{Manifold density}
    \label{fig:app-bt-scm1-manifold}
  \end{subfigure}\hfill
  \begin{subfigure}[t]{0.3\textwidth}
    \centering
    \includegraphics[width=\linewidth]{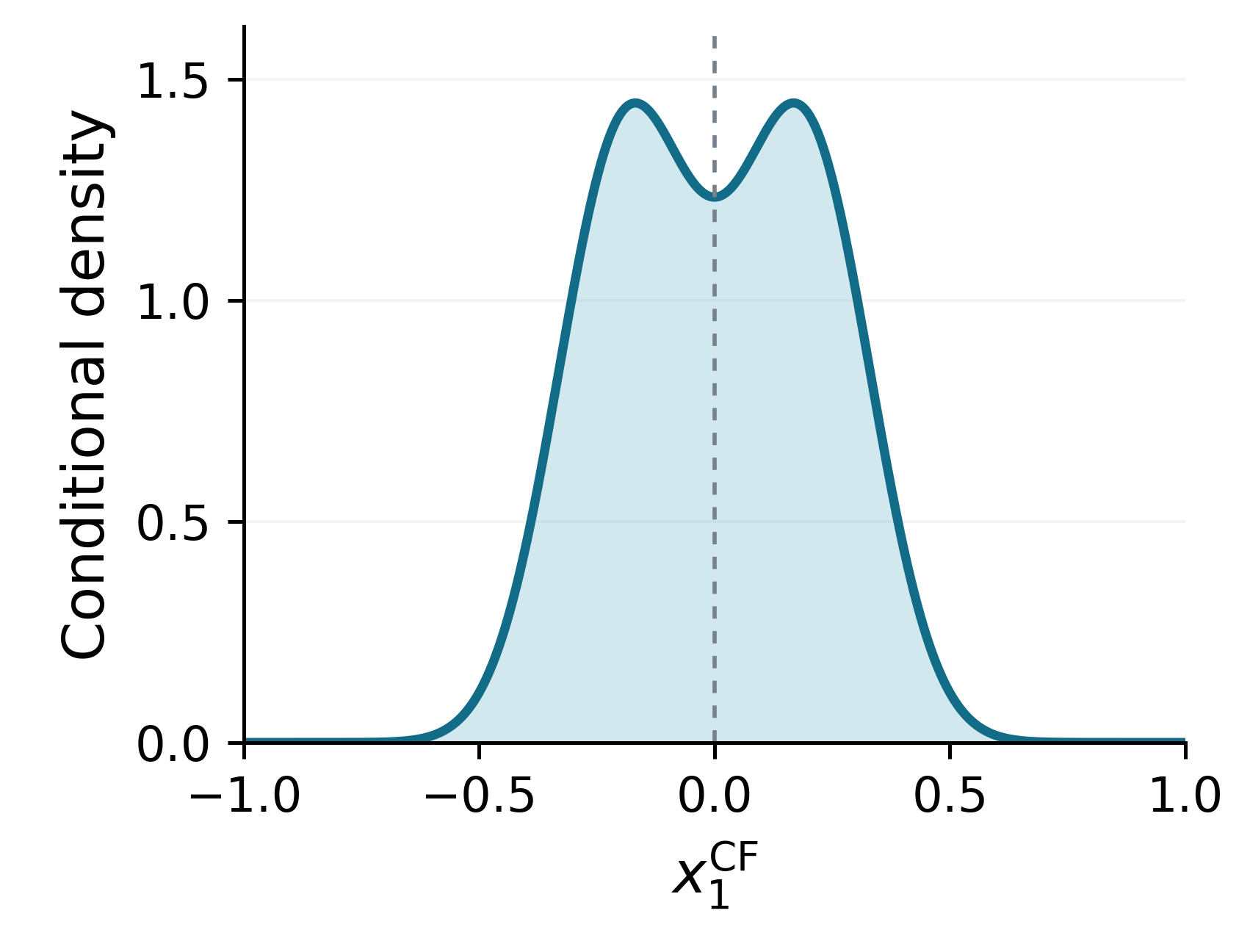}
    \caption{Density of $X_1^{\mathrm{CF}}$}
    \label{fig:app-bt-scm1-x1}
  \end{subfigure}\hfill
  \begin{subfigure}[t]{0.3\textwidth}
    \centering
    \includegraphics[width=\linewidth]{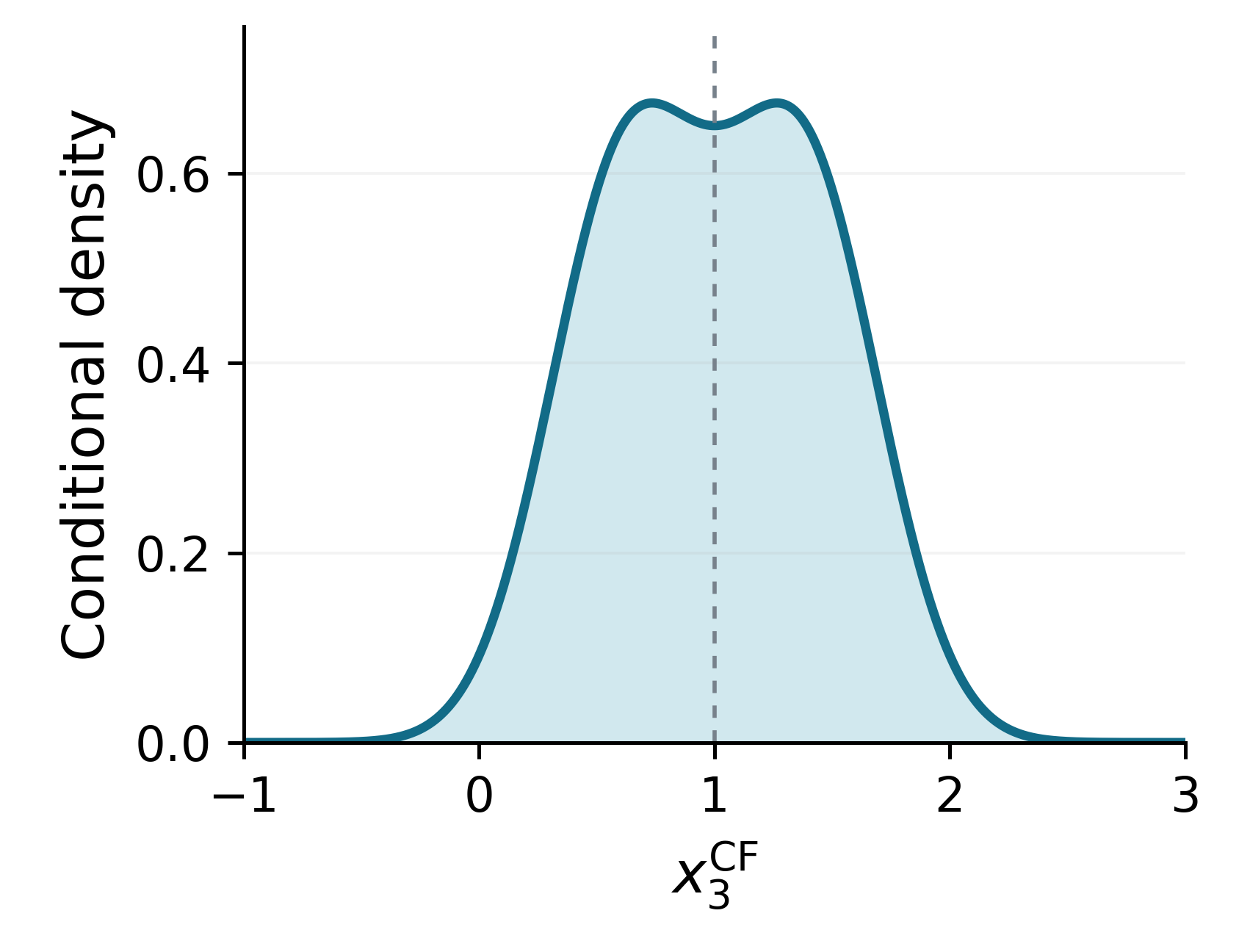}
    \caption{Density of $X_3^{\mathrm{CF}}$}
    \label{fig:app-bt-scm1-x3}
  \end{subfigure}
  \caption{
  Hard-constraint backtracking counterfactuals for the SCM
  $x_1=u_1$, $x_2=x_1^2+x_1u_2+u_2^2$, $x_3=x_1x_2+u_3$.
  The observation is $\mathbf{x}=(0,1,1)$, the counterfactual constraint is
  $x_2^{\mathrm{CF}}=2$, and the prior is $\mathcal{N}(0,I)$.
  }
  \label{fig:app-bt-scm1}
\end{figure*}

\begin{figure*}[htbp]
  \centering
  \begin{subfigure}[t]{0.3\textwidth}
    \centering
    \includegraphics[width=\linewidth]{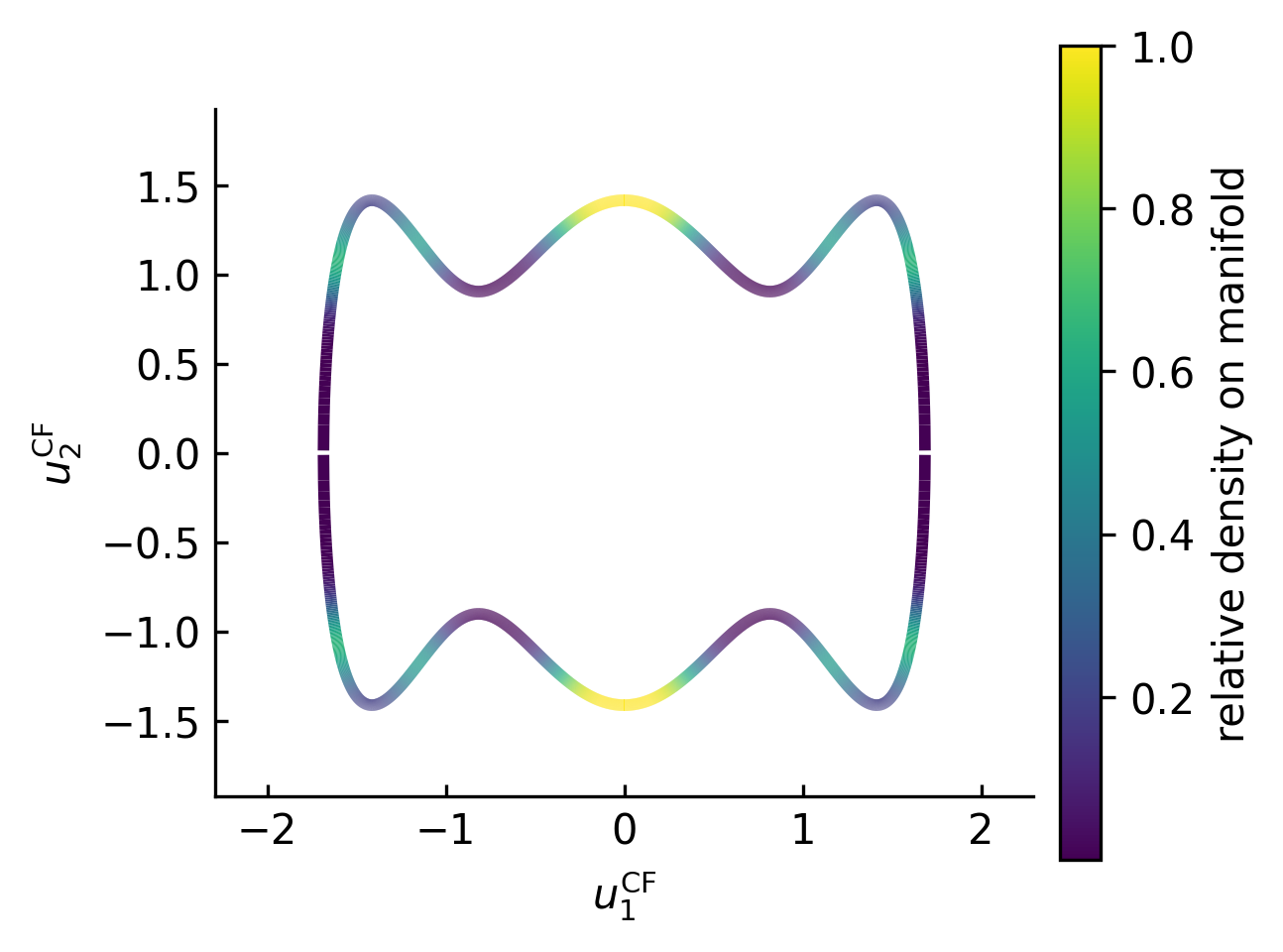}
    \caption{Manifold density}
    \label{fig:app-bt-scm2-manifold}
  \end{subfigure}\hfill
  \begin{subfigure}[t]{0.3\textwidth}
    \centering
    \includegraphics[width=\linewidth]{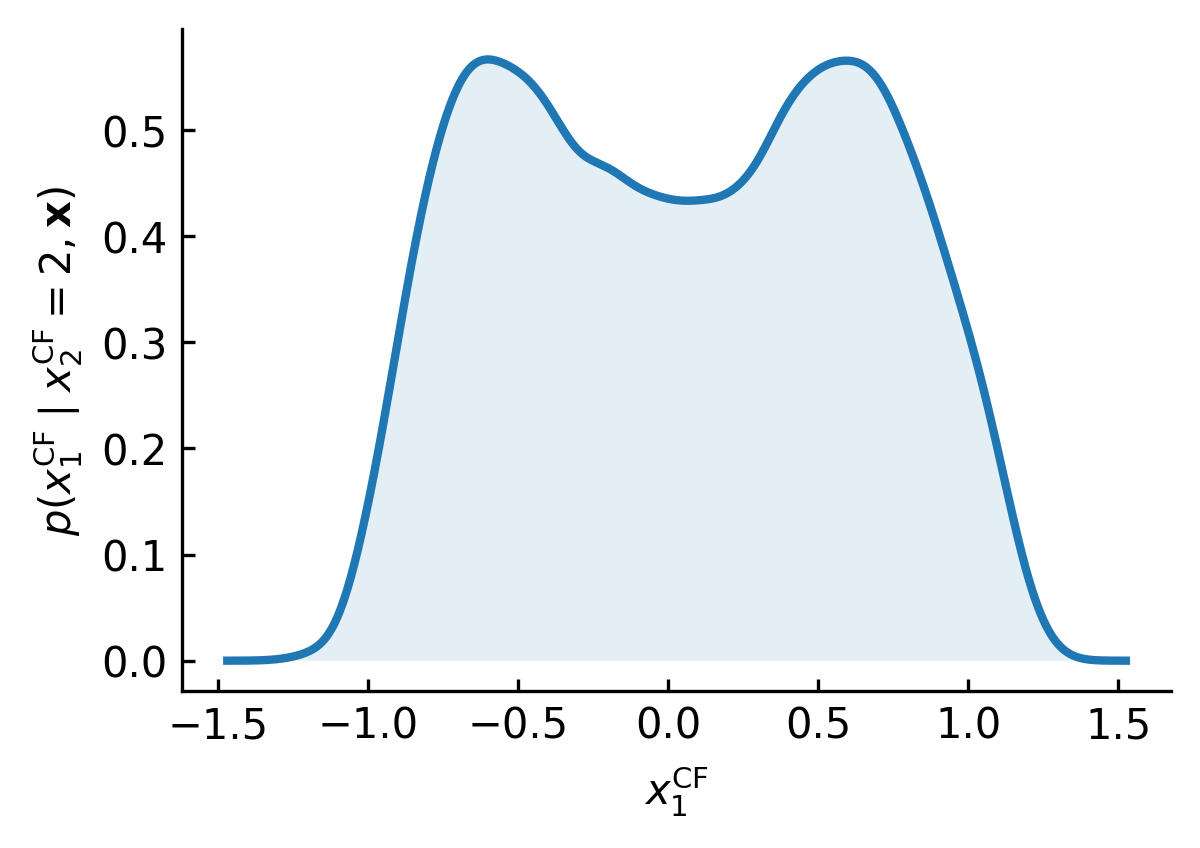}
    \caption{Density of $X_1^{\mathrm{CF}}$}
    \label{fig:app-bt-scm2-x1}
  \end{subfigure}\hfill
  \begin{subfigure}[t]{0.3\textwidth}
    \centering
    \includegraphics[width=\linewidth]{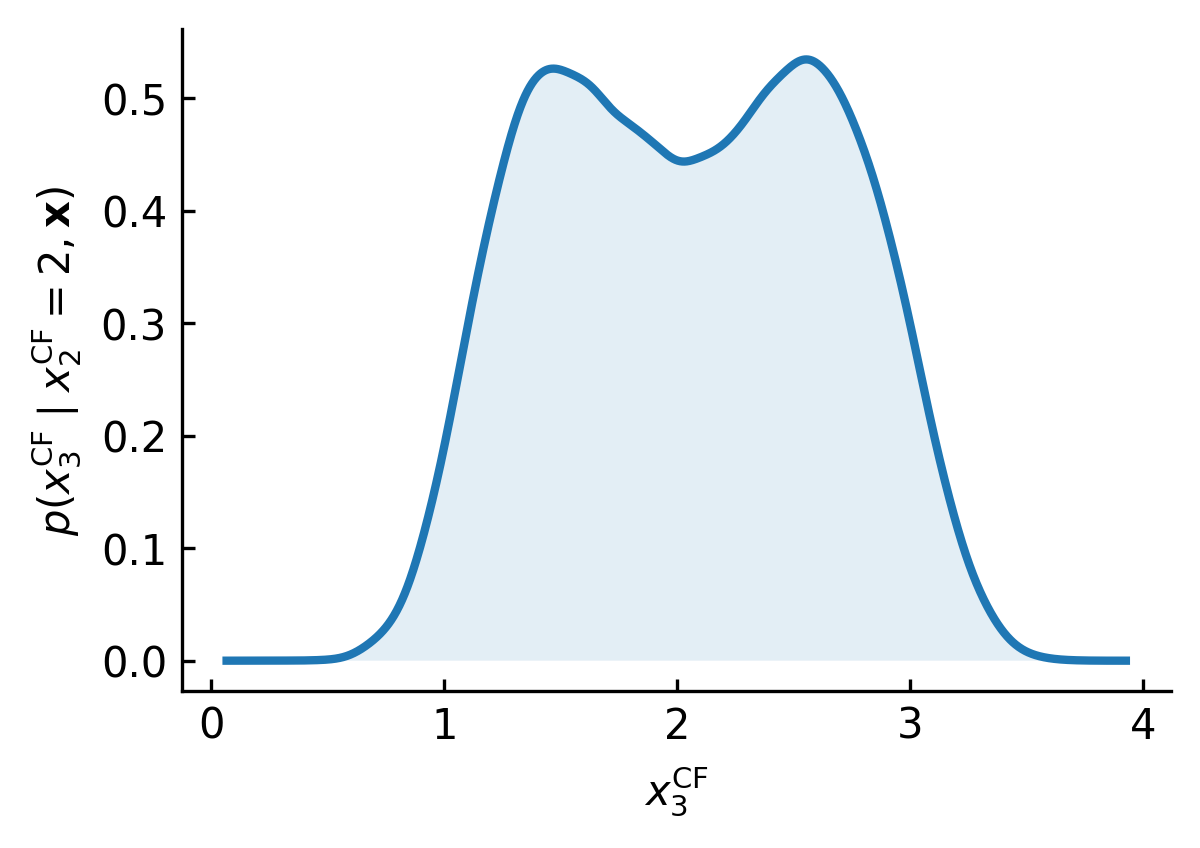}
    \caption{Density of $X_3^{\mathrm{CF}}$}
    \label{fig:app-bt-scm2-x3}
  \end{subfigure}
  \caption{
  Hard-constraint backtracking counterfactuals for the SCM
  $x_1=f(u_1)=u_1^3-2u_1$, $x_2=x_1^2+u_2^2$, $x_3=x_1+x_2+u_3$.
  The observation is $\mathbf{x}=(0,1,1)$, the counterfactual constraint is
  $x_2^{\mathrm{CF}}=2$, and the prior is $\mathcal{N}(0,I)$.
  }
  \label{fig:app-bt-scm2}
\end{figure*}

\Cref{fig:app-bt-scm1,fig:app-bt-scm2} show density-level
visualizations for hard-constraint backtracking counterfactuals. These figures
use the cross-world kernel
$p(u^{\mathrm{CF}}\mid u)\propto\exp(-\lambda\|u^{\mathrm{CF}}-u\|_2^2)$ with
$\lambda=20$.

\subsection{Synthetic SCMs and Detailed Results}
\label[appendix]{app:synthetic-full}

\Cref{tab:synthetic-summary} provides the aggregate comparison discussed in
\Cref{subsec:synthetic-experiment-main}. The individual SCM definitions
and per-instance results follow.

\paragraph{Five-dimensional nonlinear SCM.}
\[
    \begin{cases}
        X_1 = \sin(U_1),\\
        X_2 = \exp(X_1) + |U_2|,\\
        X_3 = \max\{X_1, X_2\} + \cos(U_3),\\
        X_4 = \log(1 + |X_1|) + X_2 + U_4^2,\\
        X_5 = X_4 + \tanh(U_5) + \min\{X_2, X_3\}.
    \end{cases}
\]
The observation is $x_{\mathrm{obs}}=(0.0,2.0,3.0,3.0,5.0)$ and the
counterfactual query is $x_2^{\mathrm{CF}}=1.5$, $x_4^{\mathrm{CF}}=3.5$.

\begin{table}[ht]
\centering
\caption{Results for the 5D nonlinear SCM. The common protocol is specified in \Cref{tab:solver-hyperparams-app}.}
\label{tab:app-scm5-results}
\begin{tabular}{lrrr}
\toprule
Method & $D_U$ & $L_{\mathrm{CF}}$ & $L_{\mathrm{ind}}$ \\
\midrule
\textsc{Marginal} & $0.250646$ & $1.604\times 10^{-7}$ & $0$ \\
\textsc{Penalty} & $0.250813$ & $4.119\times 10^{-8}$ & $5.166\times 10^{-13}$ \\
\texttt{OT-CF} & $0.240670$ & $1.694\times 10^{-4}$ & $1.137\times 10^{-21}$ \\
\textsc{Hard} & $0.250959$ & $1.169\times 10^{-18}$ & $0$ \\
\bottomrule
\end{tabular}
\end{table}

\paragraph{Six-dimensional nonsmooth SCM.}
\[
    \begin{cases}
        X_1 = |U_1| - 1,\\
        X_2 = \max\{0, X_1\} + \cos(U_2),\\
        X_3 = \log(\exp(X_2)) + \mathbb{I}\{U_3 > 0\},\\
        X_4 = X_3 + \min\{|U_4|, 2\},\\
        X_5 = \dfrac{X_4}{1 + e^{-U_5}},\\
        X_6 = X_5 + \arctan(U_6).
    \end{cases}
\]
The observation is $x_{\mathrm{obs}}=(1.0,2.0,3.0,5.0,2.5,2.5)$ and the
counterfactual query is $x_1^{\mathrm{CF}}=2.0$, $x_5^{\mathrm{CF}}=4.0$.

The numerical posterior uses the two signs of $U_1$, the three branches
$U_2\in\{-2\pi,0,2\pi\}$ with normalized Gaussian density weights, eight
conditional-quantile midpoint nodes for each of $U_3$ and $U_4$, and
$U_5=U_6=0$. All four Adam solvers use this same $384$-point product source
and the common protocol in \Cref{tab:solver-hyperparams-app}.

\begin{table}[ht]
\centering
\caption{Results for the 6D nonsmooth SCM. The common protocol is specified in \Cref{tab:solver-hyperparams-app}.}
\label{tab:app-scm6-nonsmooth-results}
\begin{tabular}{lrrr}
\toprule
Method & $D_U$ & $L_{\mathrm{CF}}$ & $L_{\mathrm{ind}}$ \\
\midrule
\textsc{Marginal} & $1.479059$ & $6.977\times 10^{-7}$ & $0$ \\
\textsc{Penalty} & $1.479755$ & $1.749\times 10^{-7}$ & $8.480\times 10^{-20}$ \\
\texttt{OT-CF} & $1.435321$ & $7.449\times 10^{-4}$ & $2.273\times 10^{-18}$ \\
\textsc{Hard} & $1.480453$ & $1.209\times 10^{-18}$ & $0$ \\
\bottomrule
\end{tabular}
\end{table}

The exact reference is attained by
$T(u)=(\tfrac32u_1,u_2,u_3,u_4,\log 2,u_6)$, which preserves the product
posterior structure and gives $X^{\mathrm{CF}}=(2,3,4,6,4,4)$.
Every factual posterior point has $|u_1|=2$ and $u_5=0$.
Any feasible target $v$ must satisfy $|v_1|=3$; moreover,
$X_4^{\mathrm{CF}}\leq6$ and $X_5^{\mathrm{CF}}=4$ imply $v_5\geq\log2$.
Thus $\|v-u\|_2^2\geq1+(\log2)^2$, and $T$ attains this bound for every
posterior point. Consequently, the exact OT-CF and IOT-CF optimal values
both equal $1+(\log2)^2 \simeq 1.480453.$

\paragraph{Six-dimensional polynomial SCM.}
\[
\begin{cases}
X_1 = U_1^3 - 2U_1,\\
X_2 = X_1 + U_2^4 - 5U_2^2,\\
X_3 = X_1X_2 + U_3^2,\\
X_4 = X_2^2 + X_3U_4 + U_4^2,\\
X_5 = X_3 + X_4 + U_5^2,\\
X_6 = X_1X_4 + X_5 + U_6^2.
\end{cases}
\]
The observation is $x_{\mathrm{obs}}=(1,-3,-2,9,8,18)$ and the counterfactual
query is $x_3^{\mathrm{CF}}=1$, $x_5^{\mathrm{CF}}=10$.

\begin{table}[ht]
\centering
\caption{Adam results for the 6D polynomial SCM. The common protocol is specified in \Cref{tab:solver-hyperparams-app}.}
\label{tab:app-scm6-polynomial-results}
\begin{tabular}{lrrr}
\toprule
Method & $D_U$ & $L_{\mathrm{CF}}$ & $L_{\mathrm{ind}}$ \\
\midrule
\textsc{Marginal} & $0.783031$ & $2.427\times 10^{-7}$ & $0$ \\
\textsc{Penalty} & $0.384916$ & $8.635\times 10^{-6}$ & $1.353\times 10^{-3}$ \\
\texttt{OT-CF} & $0.286562$ & $8.432\times 10^{-6}$ & $2.669\times 10^{-2}$ \\
\textsc{Hard} & $0.714376$ & $2.176\times 10^{-19}$ & $0$ \\
\bottomrule
\end{tabular}
\end{table}

\paragraph{Ten-dimensional nonlinear SCM.}
\[
\begin{cases}
X_1 = U_1^3 - 2U_1,\\[2mm]
X_2 = \log(1+X_1^2) + \cos(U_2),\\[2mm]
X_3 = e^{0.3X_1} + \dfrac{X_2}{1+X_1^2} + \bigl(U_3 + 0.2\sin(U_3)\bigr),\\[3mm]
X_4 = X_1X_2 + \log(1+U_4^2),\\[2mm]
X_5 = \tanh(X_2+X_3) + \bigl(U_5 + 0.1\sin(U_5)\bigr),\\[3mm]
X_6 = \dfrac{X_3^2}{1+X_4^2} + \sin(U_6),\\[3mm]
X_7 = e^{-0.2X_4^2} + X_5X_6 + \bigl(U_7 + 0.15\sin(U_7)\bigr),\\[3mm]
X_8 = \log(1+X_6^2) + \dfrac{X_7}{1+X_5^2} + \cos(U_8),\\[3mm]
X_9 = \tanh(X_7+X_8) + \dfrac{X_3}{1+X_2^2} + \bigl(U_9 + 0.1\sin(U_9)\bigr),\\[3mm]
X_{10} = X_8X_9 + \log(1+X_6^2) + \sin(U_{10}).
\end{cases}
\]
The observation is
\[
\begin{aligned}
x_1 &= -1.0, & x_2 &= 1.2, & x_3 &= 1.9, & x_4 &= -0.5, & x_5 &= 0.55,\\
x_6 &= 3.5,  & x_7 &= 3.6, & x_8 &= 4.8, & x_9 &= 1.0, & x_{10} &= 8.0,
\end{aligned}
\]
and the counterfactual query is $x_8^{\mathrm{CF}}=10$.

\begin{table}[ht]
\centering
\caption{Results for the 10D nonlinear SCM. The common protocol is specified in \Cref{tab:solver-hyperparams-app}.}
\label{tab:app-scm10-results}
\begin{tabular}{lrrr}
\toprule
Method & $D_U$ & $L_{\mathrm{CF}}$ & $L_{\mathrm{ind}}$ \\
\midrule
\textsc{Marginal} & $1.246304$ & $1.239\times 10^{-7}$ & $0$ \\
\textsc{Penalty} & $1.250650$ & $9.216\times 10^{-8}$ & $2.264\times 10^{-6}$ \\
\texttt{OT-CF} & $0.795014$ & $3.107\times 10^{-5}$ & $7.270\times 10^{-4}$ \\
\textsc{Hard} & $0.798030$ & $7.776\times 10^{-22}$ & $0$ \\
\bottomrule
\end{tabular}
\end{table}

\subsection{Comparison with Interventional Counterfactuals}
\label[appendix]{app:interventional-comparison}

We extend the evaluation in \Cref{subsec:additional-comparisons-main}
using the 5D nonlinear, 6D polynomial, and 10D nonlinear SCMs, with the factual
observations and counterfactual queries specified in
\Cref{app:synthetic-full}. \Cref{tab:interventional-comparison}
reports the \texttt{OT-CF} transport cost $D_U$, the interventional
exogenous-cost diagnostic, and the feasible posterior mass. The last quantity
is the posterior probability mass for which the interventional outcomes are
compatible with the original local structural mechanisms. This compatibility
check evaluates the outcomes against the unchanged mechanisms; the
interventional procedure itself uses the modified assignments described in
\Cref{app:formal-preliminaries}.

The interventional exogenous-cost column is an auxiliary comparison diagnostic.
Standard abduction--action--prediction retains the abducted noises while
replacing the queried structural assignments. Accordingly, the diagnostic
should be read together with the feasibility column: finite values for
partially or fully incompatible instances do not certify an exact realization
of their outcomes under the original mechanisms.

\begin{table}[htbp]
	\centering
	\caption{Comparison with interventional counterfactuals. The feasibility column
		measures the compatibility of interventional outcomes with the original local
		mechanisms. The \texttt{OT-CF} costs agree with
		\Cref{tab:synthetic-summary}, which also reports numerical CF residuals. A dashed line means the cost isn't applicable since it is not feasible.}
	\label{tab:interventional-comparison}
	\small
	\setlength{\tabcolsep}{7pt}
	\renewcommand{\arraystretch}{1.15}
	\begin{tabular}{@{}lrrr@{}}
		\toprule
		\textbf{Instance}
		& \textbf{\shortstack{\texttt{OT-CF}\\($D_U$)}}
		& \textbf{\shortstack{Interventional\\exogenous cost}}
		& \textbf{\shortstack{Feasible\\posterior mass for interventional}} \\
		\midrule
		5D nonlinear   & $0.2407$ & $0.4216$ & $100\%$ \\
		6D polynomial  & $0.2866$ & -- & $88.08\%$ \\
		10D nonlinear  & $0.7950$ & -- & $0\%$ \\
		\bottomrule
	\end{tabular}
\end{table}

\paragraph{Cost and mechanism compatibility.}
For the 5D nonlinear instance, the full posterior mass passes the compatibility
check, and the reported \texttt{OT-CF} cost is $42.9\%$ lower than the
interventional diagnostic, since
$(0.4216-0.2407)/0.4216\approx0.429$. For the 6D polynomial instance, only
$88.08\%$ of posterior mass passes this check, indicating that some
interventional outcomes cannot be generated by the original local mechanisms.
For the 10D nonlinear instance, the compatible mass is zero because the target
lies outside the local mechanism's range at the parent configuration retained
by the intervention. Backtracking can also change upstream exogenous variables
and hence the parent configuration. Thus, the local incompatibility of an
interventional outcome does not imply that the backtracking feasible set is
empty. These comparisons characterize the behavior of the constructions on
the tested SCMs and do not establish a universal cost advantage.

\subsection{Transfer-Kernel Sensitivity}
\label[appendix]{app:transfer-kernel-sensitivity}

We evaluate the sensitivity of probabilistic backtracking to the chosen
cross-world transfer kernel on Example~1 and Example~2. Their structural
equations are given in \Cref{app:illustrative-full}. Within each
experiment, we hold the SCM, factual observation, and counterfactual antecedent
fixed and vary only the kernel family or scale. \Cref{tab:transfer-kernel-sensitivity}
reports the transport distance $D_K$, the dependence score among transported
exogenous variables, and the standard deviation of the counterfactual outcome.

\paragraph{Kernel families and scales.}
The uniform kernel uses $h=\sqrt{3/2}$ and the Laplace kernel uses $b=0.5$;
both have coordinate variance $0.5$. The Gaussian transfer density is
\[
    k_{\lambda}(\mathbf u^{\mathrm{CF}}\mid\mathbf u)
    \propto
    \exp\!\left(-\lambda
    \|\mathbf u^{\mathrm{CF}}-\mathbf u\|_2^2\right),
\]
with coordinate variance $1/(2\lambda)$. Here $\lambda$ is the Gaussian kernel
concentration parameter. The Gaussian kernel with $\lambda=1$ therefore
matches the coordinate variance of the uniform and Laplace kernels, while
the remaining Gaussian settings examine scale sensitivity.

\begin{table}[htbp]
\centering
\caption{Transfer-kernel sensitivity on the two illustrative SCMs. For each
example, columns report transport distance $D_K$, exogenous dependence, and
counterfactual outcome standard deviation. Uniform, Laplace, and Gaussian
($\lambda=1$) have coordinate variance $0.5$; the other Gaussian rows vary
the kernel scale.}
\label{tab:transfer-kernel-sensitivity}
\footnotesize
\setlength{\tabcolsep}{4pt}
\renewcommand{\arraystretch}{1.15}
\begin{tabular}{@{}lrrrrrr@{}}
\toprule
& \multicolumn{3}{c}{\textbf{Example 1}}
& \multicolumn{3}{c}{\textbf{Example 2}} \\
\cmidrule(lr){2-4}\cmidrule(l){5-7}
\textbf{Kernel}
& $D_K$ & \textbf{Dependence} & \textbf{\shortstack{Outcome\\std.}}
& $D_K$ & \textbf{Dependence} & \textbf{\shortstack{Outcome\\std.}} \\
\midrule
Uniform ($h=\sqrt{3/2}$) & $1.275334$ & $0.355$ & $1.647$ & $1.035944$ & $0.071$ & $1.125$ \\
Laplace ($b=0.5$)       & $1.242350$ & $0.235$ & $1.554$ & $1.053674$ & $0.069$ & $1.089$ \\
\addlinespace
Gaussian ($\lambda=0.5$)  & $2.566089$ & $0.323$ & $2.201$ & $2.212211$ & $0.075$ & $1.314$ \\
Gaussian ($\lambda=1$)    & $1.353478$ & $0.266$ & $1.698$ & $1.061733$ & $0.075$ & $1.098$ \\
Gaussian ($\lambda=2$)    & $0.679153$ & $0.248$ & $1.154$ & $0.575082$ & $0.072$ & $0.956$ \\
Gaussian ($\lambda=5$)    & $0.336017$ & $0.421$ & $0.744$ & $0.301044$ & $0.068$ & $0.814$ \\
Gaussian ($\lambda=10$)   & $0.232855$ & $0.656$ & $0.584$ & $0.203912$ & $0.064$ & $0.751$ \\
Gaussian ($\lambda=20$)   & $0.182163$ & $0.785$ & $0.489$ & $0.140795$ & $0.066$ & $0.775$ \\
Gaussian ($\lambda=50$)   & $0.151979$ & $0.918$ & $0.423$ & $0.070130$ & $0.025$ & $0.915$ \\
\bottomrule
\end{tabular}
\end{table}

\paragraph{Kernel-family sensitivity at matched variance.}
Changing the kernel family affects all three reported quantities even when
coordinate variance is fixed. In Example~1, uniform, Laplace, and Gaussian
($\lambda=1$) kernels yield dependence scores of $0.355$, $0.235$, and $0.266$,
respectively, and outcome standard deviations of $1.647$, $1.554$, and $1.698$.
The same family comparison also changes the reported quantities in Example~2.

\paragraph{Gaussian scale sensitivity.}
Across the tested Gaussian parameters, increasing $\lambda$ from $0.5$ to $50$
reduces $D_K$ from $2.566089$ to $0.151979$ in Example~1 and from $2.212211$
to $0.070130$ in Example~2. The dependence score and outcome dispersion need
not follow a common trend. In Example~1, dependence first falls to $0.248$
at $\lambda=2$ and then rises to $0.918$ at $\lambda=50$, while outcome
standard deviation decreases from $2.201$ to $0.423$. In Example~2, outcome
standard deviation decreases from $1.314$ to $0.751$ at $\lambda=10$ and
then increases to $0.915$ at $\lambda=50$. Thus, concentrating the kernel
can have different effects on dependence and outcome dispersion across SCMs.

\paragraph{Implications for the transport formulation.}
The ablation shows that an externally selected kernel can materially affect
the counterfactual distribution even when the model and query are fixed.
Our variational formulation selects the distribution through the transport
objective and feasibility constraints, after which an optimal coupling
induces a valid backtracking kernel
(\Cref{thm:otcf-as-bcf-clean}). The ground cost and constraints remain
explicit modeling choices. These experiments demonstrate kernel sensitivity
in controlled settings and do not establish scalability to learned or large
real-world SCMs.

\end{document}